\documentclass[12pt,numbers]{article}

\usepackage{graphicx}
\usepackage{authblk}
\usepackage{natbib}

\usepackage{amsfonts}

\usepackage{amsmath}
\usepackage{url}
\usepackage{tabularx}
\usepackage{multirow}
\usepackage{booktabs}
\usepackage{subfigure}

\usepackage[normalem]{ulem}

\begin{document}

\title{Analysis of the Robustness of an Edge Detector Based on Cellular Automata Optimized by Particle Swarm}

\author[ ]{Vin\'{i}cius Ferraria$^{2}$}
\author[ ]{Eurico Ruivo$^{1,2}$}

\affil[1]{Faculdade de Computa\c{c}\~{a}o e Inform\'{a}tica}
\affil[2]{P\'{o}s-Gradua\c{c}\~{a}o em Engenharia El\'{e}trica e Computa\c{c}\~{a}o}
\affil[ ]{Universidade Presbiteriana Mackenzie}
\affil[ ]{Rua da Consola\c{c}\~{a}o 896, Consola\c{c}\~{a}o -- 01302-907, S\~{a}o Paulo, SP,  Brazil}
\affil[ ]{\texttt{\small vferraria10@gmail.com,eurico.ruivo@mackenzie.br}}

\date{}
\maketitle

\begin{abstract}
The edge detection task is essential in image processing aiming to extract relevant information from an image. One recurring problem in this task is the weaknesses found in some detectors, such as the difficulty in detecting loose edges and the lack of context to extract relevant information from specific problems. To address these weaknesses and adapt the detector to the properties of an image, an adaptable detector described by two-dimensional cellular automaton and optimized by meta-heuristic combined with transfer learning techniques was developed. This study aims to analyze the impact of expanding the search space of the optimization phase and the robustness of the adaptability of the detector in identifying edges of a set of natural images and specialized subsets extracted from the same image set. The results obtained prove that expanding the search space of the optimization phase was not effective for the chosen image set. The study also analyzed the adaptability of the model through a series of experiments and validation techniques and found that, regardless of the validation, the model was able to adapt to the input and the transfer learning techniques applied to the model showed no significant improvements.

{\it Keywords}: cellular automaton, natural computing, edge detection, image processing, transfer learning

\end{abstract}

\section{Introduction}
The edge detection task may seem simple, however, its role is of paramount importance to the area of visual computing. In many of its applications, the edge detection task is applied as an initial step to extract relevant information from images, serving as the basis for \textit{3D} mapping applications \cite{basiri2023scalable,arora2021Mapping3DLidar}, object recognition \cite{rani2022object,yasir2022object} and image segmentation \cite{al2010image,techniquesSegmentationBasedEdgeDetection}.

Edge detectors have two classes and can be classified as autonomous detectors, those that do not use \textit{a prior} knowledge for the task, and contextual detectors, those that make use of relevant information about the objective to improve detection \cite{edgedetectionoverview1998}. Aiming to address the problem of the edge detection task, several models of edge detectors have been proposed. The first detectors developed are relatively simple and belong to the autonomous category, mapping the edges by extracting local information such as the intensity of the adjacent pixels of a given neighborhood. A common weakness of these detectors and their class is their need to generalize the task, evidenced in more complex and specific scenarios that require the detection of loose edges and the extraction of intrinsic characteristics of image categories.

With the evolution of studies in the area and the increase of computing power, new models of contextual detectors have emerged that propose to solve the problems of generalization faced by previous detectors by combining the knowledge of several other areas of research such as artificial intelligence and cellular automata. Examples of these new applications include models based on neural networks that learn from previously evaluated data \cite{wang2016edge, pu2022edter} and the use of the properties of cellular automata \cite{mohammed2014efficient} that increase the processing capacity and viability of contextual edge detectors.

The use of cellular automaton to solve problems, from the simplest to the most complex, is a topic widely covered in the literature. This interest is the result of its discrete and intrinsically parallel nature, which uses only local information to achieve a global objective, making them conducive to the development of the most diverse types of systems such as circuits \cite{khan1997vlsi,adder2007qca,torres2018qca}, real event simulators \cite{sante2010cellular}, processing of computational tasks \cite{dumitru2021transfer}, simulation of panic contagion in a dynamic population \cite{wang2022simulation}, modeling and geosimulation of global changes in the use of urban lands \cite{addae2022integrating} and the real-time modeling of evacuation systems of buildings and schools \cite{buildings12060718}. The application of cellular automaton for image processing is the result of the presence of extremely similar characteristics between both areas such as the use of a pixel neighborhood (local information) for edge detection (global objective).

Some of the implementations of edge detectors based on cellular automaton can be described by the application of linear rules and the use of the Moore neighborhood \cite{mohammed2014efficient}. The use of the rules and topology of cellular automaton opened a range of possibilities due to the diverse types of behaviors presented by the composition of each rule, the radius of its vicinity, and the states of each cell. In order to explore and benefit from all the properties and behaviors of cellular automaton, models have been developed that combine its application with meta-heuristics, exploration techniques based on and inspired by various phenomena of nature, in order to explore the space of possible solutions and find the optimal solution, that is, the solution that presents the best rule composition and topology of an automaton given a fitness function considering the specific characteristics of each image \cite{dumitru2020evolutionary,dumitru2020robustness,dumitru2021transfer}.

The objective of this research is to quantitatively analyze the robustness and generalization of an adaptive edge detector model based on gray-scale images \cite{dumitru2021transfer} that was originally developed for problems related to medical images, which makes use of a two-dimensional cellular automaton and is optimized by the Particle Swarm Optimization (PSO) in combination with Transfer Learning techniques.

\section{Materials and Methods}
\subsection {Edge Detection Task}
Edge detection is the process of detecting the edges of an image. An edge is detected when the difference between a pixel and its neighborhood is considered abrupt or significant; that is, it is higher than a previously defined threshold \cite{edgedetectionoverview1998}.

An edge detector has as input an image, preferably in black and white, and it performs the analysis of the image in its entirety using the threshold criterion mentioned above. At the end of its execution, a binary edge map is generated containing information about all the edges found.

The edge detection process is the result of the application of three operations: smoothing, differentiation, and edge classification \cite{edgedetection1995book}. Edge detectors are classified \cite{edgedetectionoverview1998} by the way they are trained:

\begin{itemize}
    \item autonomous detectors are those that use only local information such as the values of their pixel neighborhood
    \item contextual detectors are those that use some a \textit{priori} knowledge of the task during the detection process
\end{itemize}

The Canny edge detector is a general benchmark detector that serves as a metric for the development of new models. Canny adheres to three optimal criteria for the detection task: a good detection of the edges in the image, the accurate localization of the edges, and a single response to an edge.

Canny can be summarized in the following steps: the application of the Gaussian filter in the original image for the reduction of noise and calculation of the intensity of the gradients, and then the detection of maximum points of the gradient of the image.

From this detection, such maximum points go through a double threshold suppression function; values above the thresholds are classified as strong edges, values that are between the thresholds are classified as weak edges, and smaller values are discarded. After the classification of the maximums, the map is generated by selecting only the strong edges and the weak edges nearby or connected to strong edges \cite{canny1986computational}.

\subsection{Two-Dimensional Cellular Automaton}
A cellular automaton (CA) is a state machine that operates by applying a set of state transition rules to a cell based on its input, its previous states, and its neighborhood \cite{sayama2015introduction}. The cells of an automaton evolve over time from an initial configuration defined by the initial state of the cells, and each period consists of the application of the transition rules, considering the current state of the cell and its neighbors.

A two-dimensional cellular automaton has a topology of an infinite rectangular grid. Each cell of the grid is represented by an element of a two-dimensional grid referenced by $\mathbb{Z}^2$ \cite{KARI20053}. More precisely, a two-dimensional CA is given by a triple $(S,N,f)$, in which:
\begin{itemize}
    \item $S=\{0,1,\cdots,k-1\}$ is its \emph{state set};
    \item $N = (\vec{v_1}, \vec{v_2}, \ldots, \vec{v}_n)$, with $\vec{v}_i \in \mathbb{Z}^2,1\leq i\leq n$, is its \emph{neighborhood vector};
    \item $f: S^n \rightarrow S$ is the transition rule or local function that receives an input consisting of $n$ states and outputs a single state.
\end{itemize}

\begin{figure}[!ht]
\label{fig:neighborhoods}
\centering
    \subfigure[\label{fig:ca_moore}]{\includegraphics[height=1.2in,keepaspectratio=true,scale=1]{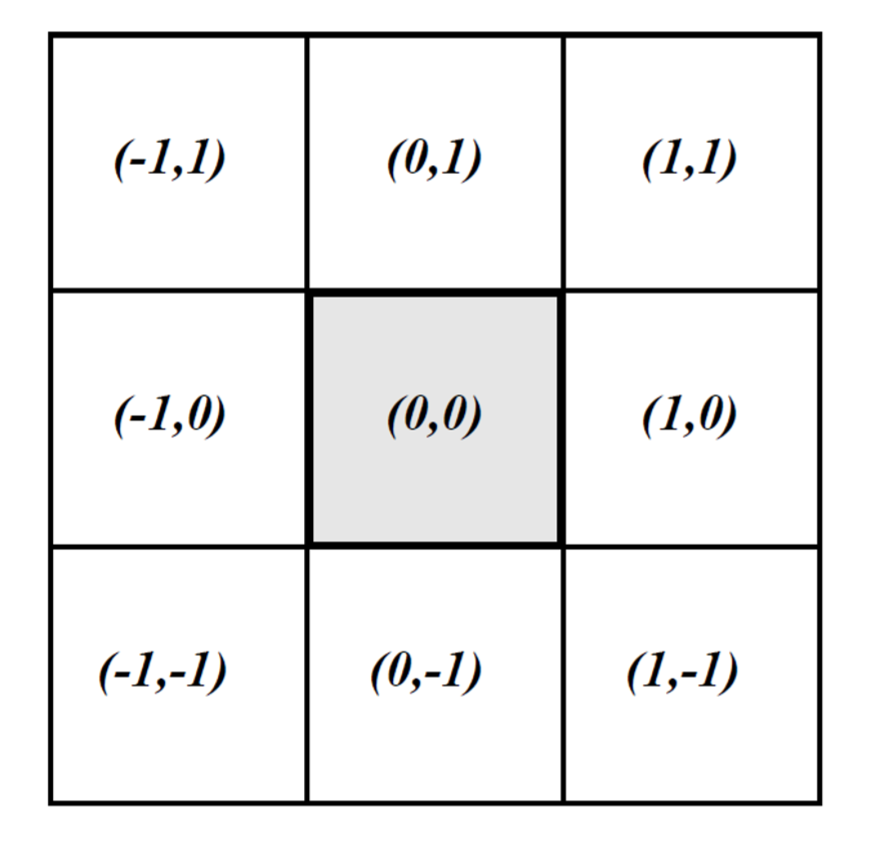}}
    \subfigure[\label{fig:ca_von_neumann}]{\includegraphics[height=1.2in,keepaspectratio=true,scale=1]
    {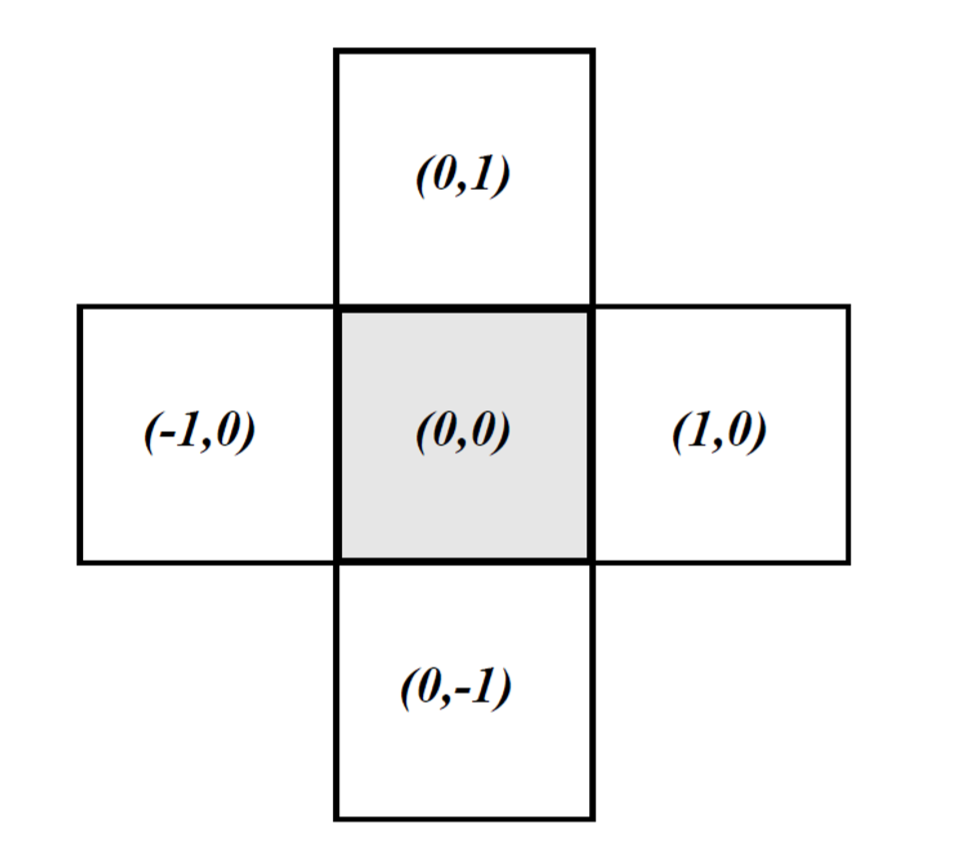}}

    \subfigure[\label{fig:ca_moore_r2}]{\includegraphics[height=1.2in,keepaspectratio=true,scale=1]{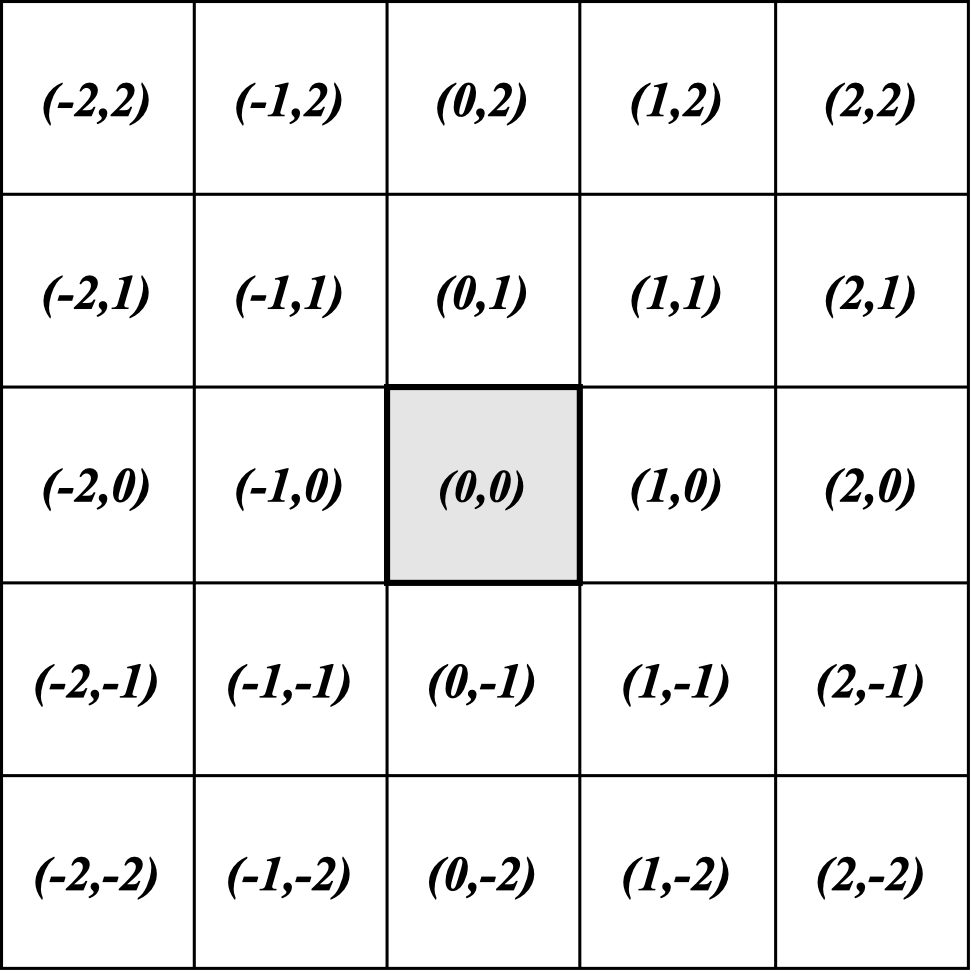}}
    \subfigure[\label{fig:ca_von_neumann_r2}]{\includegraphics[height=1.2in,keepaspectratio=true,scale=1]{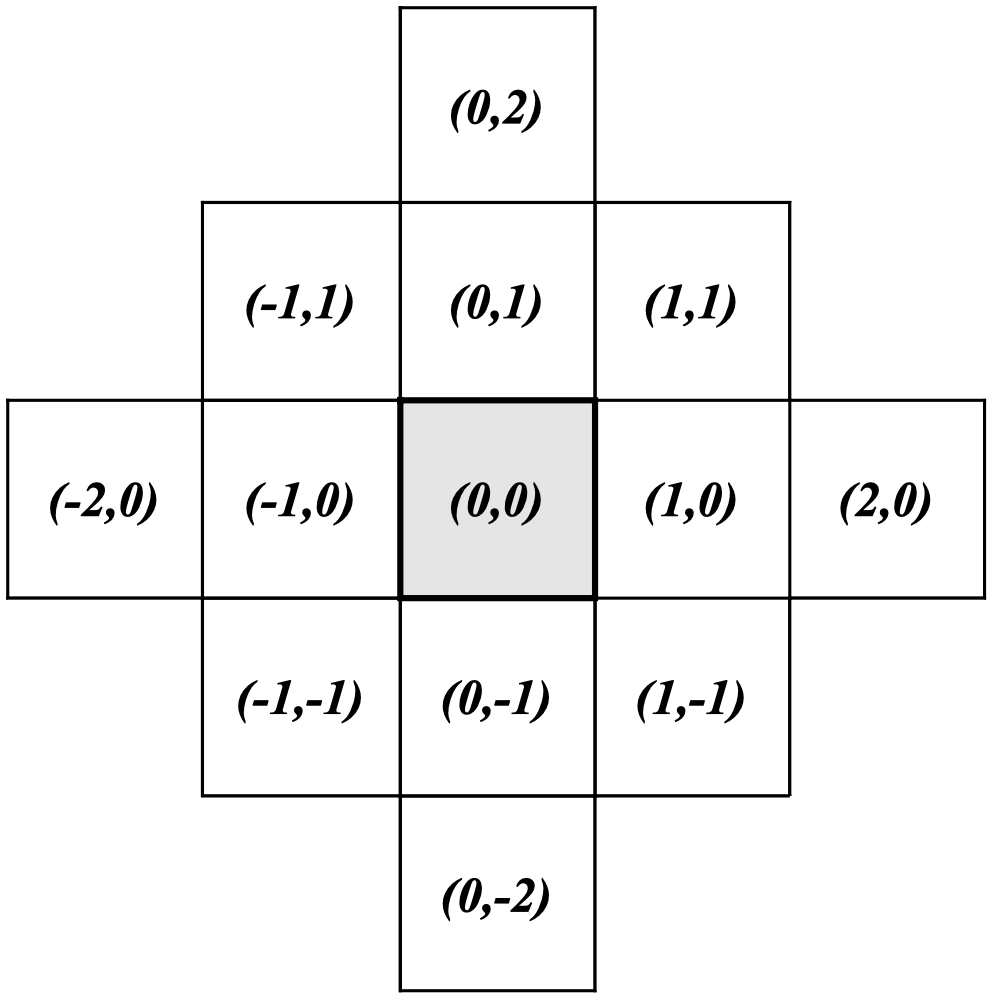}}
    
    \caption{von Neumann (left) and Moore (right) neighborhoods of radii 1 \cite{dumitru2021transfer} (top) and 2 (bottom).}
\end{figure}

 The typical neighborhood vectors for two-dimensional CAs are the von Neumann and the Moore neighborhood of radius $r\geq1$. Figure ~\ref{fig:neighborhoods} depicts such neighborhoods for radii 1 and 2.  The Moore neighborhood can be seen as an extension of the respective von Neumann neighborhood by including the neighbors in the diagonal positions to the central cell.

\subsection{Cellular Automaton Edge Detector}
This study is inspired by \cite{dumitru2020robustness} and \cite{dumitru2020evolutionary}. The contextual edge detector model presented in the reference work is described by a two-dimensional cellular automaton with a Moore neighborhood of $r = 1$ optimized by Particle Swarm Optimization for the task of edge detection in unannotated medical exam images, applying batch processing techniques and concepts of Transfer Learning, which consist of training the model on a generalized problem and then applying it to a related problem \cite{weiss2016survey}. The proposed model utilizes transition rules defined by \textit{XOR} operations between \cite{khan1997vlsi} cells and employs the representation in which a pixel \textit{X} positioned at (\textit{i}, \textit{j}) in the input image corresponds to a cell of the automaton.

For this study, the edge detector model will be adapted by expanding its radius $r$ and will also be used to extract edges from different categories of images.

The studied edge detector will be referred to as \textit{PSO-CA} during the research. The \textit{PSO-CA} is represented by a matrix, $3 \times 3$ for $r = 1$ and $5 \times 5$ for $r = 2$, with binary states to extract edge maps from grayscale images. 

\begin{figure}[pb]
\centering
\includegraphics[height=1.2in,keepaspectratio=true,scale=0.7]{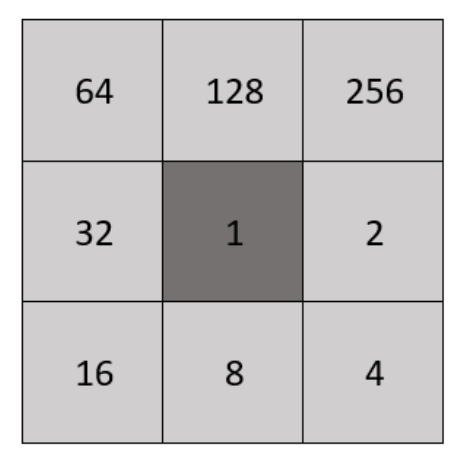}%
\caption{Representations of the proposed rule numbers of Moore's neighborhood $r = 1$ \cite{dumitru2021transfer}.}\label{fig:linear_rules_r1}
\end{figure}

The \textit{PSO-CA} defines the neighbors of a pixel positioned at \((x_{0}, y_{0})\) in a two-dimensional image as
\begin{equation}
    N(x_{0}, y_{0}) = \{(x, y): |x - x_{0}| \le r, |y - y_{0}| \le r \},
\end{equation}
where $r$ represents the radius of the neighborhood for this survey $r \in \{1, 2\}$. A visual representation of the Moore neighborhood can be seen in Figure ~\ref{fig:linear_rules_r1}. 

A two-phase transition rule defined in \cite{uguz2015edge} is used, with its first phase defined as 
\begin{equation}
    \mu_{i,j} = \frac{\phi(X_{i,j})}{\Delta + \phi (X_{i, j})},
\end{equation}
where $X_{ij}$ represents a pixel of a grayscale image $X$, a matrix with dimensions $A \times B \times 1$, in the row $i$ and column $j$. In the formula, $X_{ij}$ denotes the center cell, $\Delta \in \{0, ..., 255\}$. $\phi(X_{i,j})$ is defined as
\begin{equation}
    \phi(X_{i,j}) = \sum_{k}\sum_{q}|X_{ij} - X_{i+k,\ j+q}| ,
\end{equation}
where $k$ and $q$ are all values in the range $[-r, r]$.

The second phase of the transition rule is defined by a threshold function $F: X \mapsto \{0.1\}$ that returns the binary state, described as
\begin{equation}
    F(X_{i,j}) = \begin{cases}
    1, & if\; \mu(X_{i,j}) > \tau \\
    0, & if\; \mu(X_{i,j}) \le \tau,
    \end{cases}
\end{equation}
where $\tau\in [0,1]$.

The model has 3 hyper-parameters that need to be adjusted for each problem. $\Delta$ a parameter inversely proportional to the number of edges detected, the linear rule $z$ that defines the neighbors that the rule should consider, and the threshold $\tau$ that controls how many points are defined as edges.
The model is combined with \textit{PSO} to optimize it for the characteristics of an image set. In order to preserve the original dimensions of the images, the images will be padded by filling them with 0 for each dimension before the application of the first transition rule. The borders added to the images will result in images whose final size will be equivalent to the addition of $r$ to the $X$ and $Y$ axes of the images.

\begin{figure}[!b]
\centering
\includegraphics{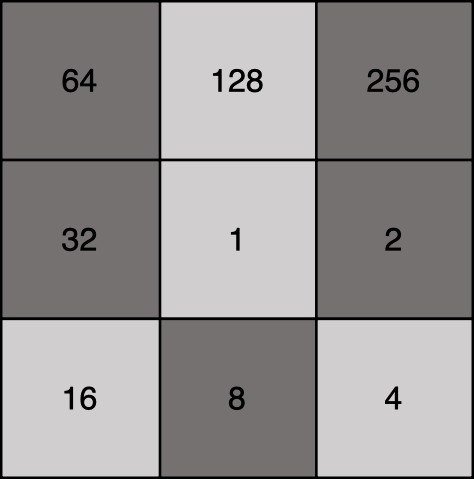}
\caption{Definition of rule 362 for $r = 1$, rule composed by the combination of multiple cells \cite{dumitru2021transfer}.}\label{fig:linear_rules_exemplo}
\end{figure}

\begin{figure}[pt]
\centering
\includegraphics{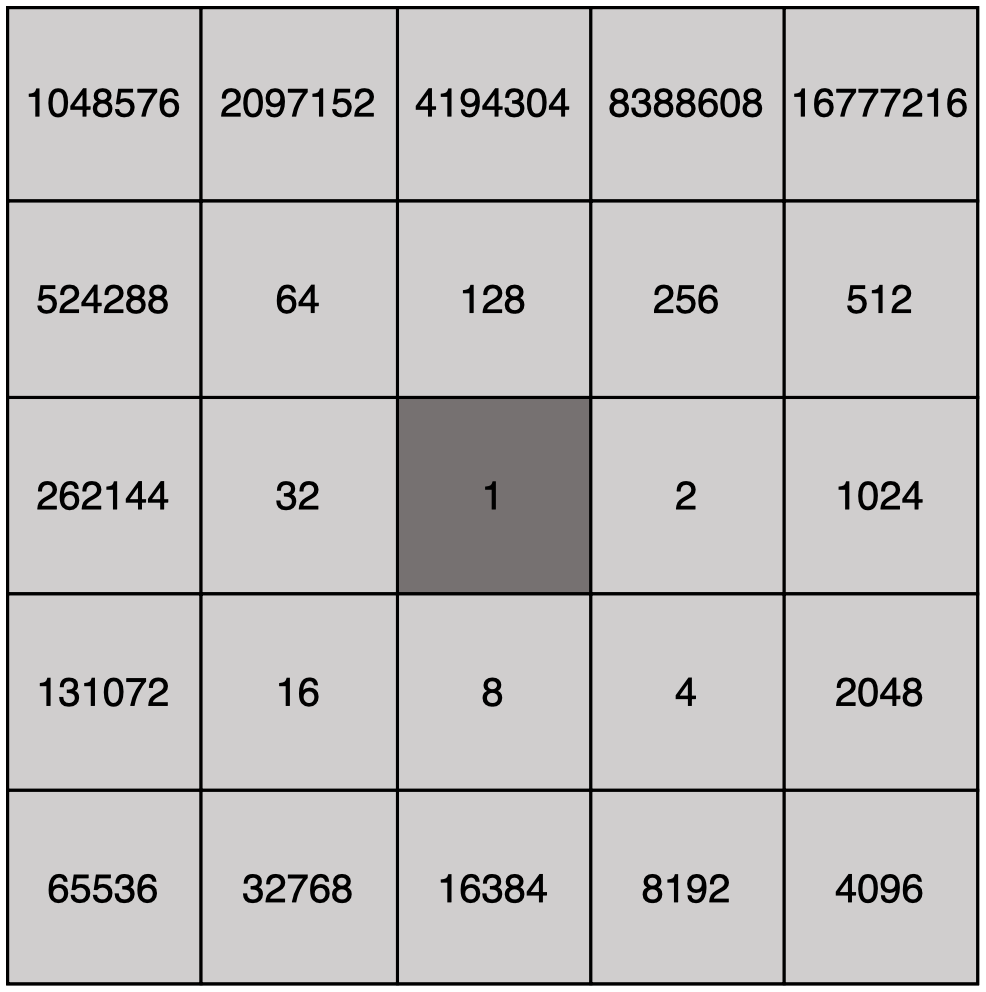}
\caption{Representations of the proposed rule numbers of Moore's neighborhood $r = 2$.}\label{fig:linear_rules_r2}
\end{figure}

The number of each transition rule is defined by a convention \cite{dumitru2021transfer}. The Figures ~\ref{fig:linear_rules_r1} and ~\ref{fig:linear_rules_r2} exemplify the numbering of the rules for the Moore neighborhoods of $r = 1$ and $r = 2$, respectively. From the definition proposed by \cite{dumitru2021transfer}, each cell of the automaton is associated with a number, and a rule is defined by the sum of the values of cells that are considered; that is, the neighborhood rule where $r \in \{1,2\}$, which considers only the central element, is rule 1. For rules that are composed of the combination of several cells, the rule number is defined by the sum of the values of the cells in consideration (Figure ~\ref{fig:linear_rules_exemplo}).

\pagebreak

\subsection{Particle Swarm Optimization}

The \textit{PSO} is a bio-inspired optimization algorithm inspired by the social behavior of a population of particles, in which each particle represents a possible solution. A particle \(i\) is defined by its position \(x_i\) in space and by its speed \(v_i\), and both influence its movement in the optimization phase \cite{kennedy1995particle}.

The PSO algorithm is executed for a specified number of iterations. During each iteration, the positions and speeds of the particles are influenced by a fitness function that assesses the solution of each particle.

Optimization is achieved through the exploration of solutions defined by the movement of particles within a search space. This movement is driven by the particles’ experiences and interactions with neighboring particles.

A particle $i$’s position at time $t$ during an iteration is determined by its speed, which is calculated using the following formula:

\begin{equation}
    V_i(t+1) = w \cdot v_i(t) + r_1 \cdot c_1 \cdot (p_{best_{i}} - x_i(t)) + r_2 \cdot c_2 \cdot (g_{best} - x_i(t))
\end{equation}
where $w$ controls the oscillation of the particle $p$, $p_{best}$ is the best position of the particle $i$, $g_{best}$ is the best global position of the swarm, $c_1$ is the coefficient referring to the particle's own influence, $c_2$ is the coefficient of the influence factor of the swarm, and $r_1,r_2 \in (0,1)$ are random uniform variables. $x_i(t)$ represents the position of the particle $i$ in time $t$ and is defined by the formula:

\begin{equation}
    x_{i}(t+1) = v_{i}(t+1) + x_i(t)
\end{equation}

The particles move through the search space along the iterations and tend to approach the position that represents the global optimum for the problem.

\subsection{Application of the Particle Swarm Optimization}
The proposed \textit{PSO} is an adaptation of the one used by the reference model. The \textit{PSO} will optimize the model parameters in order to find the global best value for the fitness function, the Dice Similarity Coefficient, defined in the metrics section. For this application, for each sample of the dataset, the objective is to optimize the triple of parameters of the detector $(\Delta,\tau, z)$ where $\Delta$ and $\tau$ are defined in the section of the proposed model and the number of the linear rule, $z \in \{0, ..., z_{r_{max}}\}$ and $z_{r_{max}}$ is equivalent to the maximum number of the rule, following the convention defined in the section of the proposed model, for the radius $r$ and can be calculated through the formula $(2 ^{(1 + 2 \times r)^2)} - 1$. To match the values to a continuous domain, the triple is normalized in the range $[0, 1]$ and the final representation is described by $(\Delta',\tau, z')$ where $z' = z/z_{r_{max}}$ and $\Delta' = \Delta/255$.

In order to generalize to the entire dataset, the proposed \textit{PSO} will also make use of batch processing techniques used by the reference model. However, as the research does not aim to analyze the impact of the batch size, it will be restricted to using all the images received as an input as a single batch, and each particle will have its value calculated as the average of the result of the fitness function applied to the batch of images.

\subsubsection{Cross-Validation in Categories}

The Cross-Validation in Categories mentioned is described by training a model using a subcategory of the dataset and then validating it in the remaining subcategories. The process is repeated until each subcategory of the dataset has been used as a training set.

\subsubsection{Dataset}
The dataset used for this research is the BSDS500 \cite{amfm_pami2011}, a set of natural colored images with hand-labeled edge maps and segmentation maps very popular in edge detection and image segmentation studies.

The BSDS500 consists of 500 images, originally separated into 200 images for training, 100 for validation, and 200 for testing. Each image has an edge map labeled from 4 to 9 people. The annotated edge maps of the set are stored in \textit{MATLAB} files and grouped by photo. For the search, the maps were extracted with the help of the library \textit{scipy} \cite{2020SciPy-NMeth} and converted to \textit{PNG}.

\section{Results}
\subsection{Experimental Setup}
Due to the nature of the images, the annotated edge maps, and the variations in annotations per annotator, it is common in the literature to generate a probability map for each image from the average of the classification among all classifiers. The probability of the edge varies between 0 and 1, being 0 for a general consensus that the pixel is not an edge and 1 that the pixel is an edge. Through the probability map, it is possible to define a threshold; edges classified with probabilities above such a threshold are considered edges, and those below are disregarded \cite{pu2022edter,liu2017richer,he2019bi}.

In addition to the extraction of the edge map of the annotations by the probability threshold, the images and annotations will go through a pre-processing phase in order to adapt them to the expected input by the model and to speed up processing, allowing all images to be uploaded in batches of processing by libraries such as \textit{pytorch} \cite{Paszke_PyTorch_An_Imperative_2019}.

\begin{figure}[h!]
    \centering
    \subfigure[\label{fig:6046_resized_emt_02}]{\includegraphics[width=0.3\textwidth]{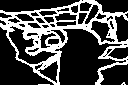}}
    \hfill
    \subfigure[\label{fig:5096_resized_emt_02}]{
        \includegraphics[width=0.3\textwidth]{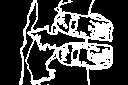}}
    \hfill
    \subfigure[
        \label{fig:2018_resized_emt_02}]
        {\includegraphics[width=0.3\textwidth]{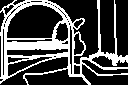}}
    \hfill
    \subfigure[\label{fig:14085_resized_emt_02}]{\includegraphics[width=0.3\textwidth]{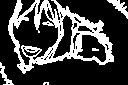}}
    \caption{
    Visualization of the pre-processed annotated edge maps of the BSDS500 dataset extracted with the probability threshold $p = 0.2$
       }
    \label{fig:bsds500_prob_map_processed}
\end{figure}

Pre-processing is described by the conversion of color images to gray scale images, followed by resizing to $128 \times 128$ maintaining the original aspect ratio and restricting the larger side dimension and standardizing the orientation of the images. Although the images of the dataset have the same final resolution, their orientations are different; some of the images have portrait orientation and some have landscape orientation. A sample of the images from the image set after the pre-processing step and their respective annotations from the edge map extracted for the probabilities 0.02 is displayed in Figure ~\ref{fig:bsds500_prob_map_processed}.

In order to explore the adaptability of the model to the intrinsic characteristics of images, the division of sets of the original dataset will be disregarded, and the dataset will go through a new stage of division in which it will be divided into subsets considering what is in focus in the image. The four categories found in the dataset were: animals, people, objects, and landscapes. The division of the 500 images between the categories found was: 195 images of animals, 100 images of landscapes, 54 images of objects, and 151 images of people.

\subsubsection{Experiments}
From the definition of the model proposed by the research, the detector was implemented using the pytorch \cite{Paszke_PyTorch_An_Imperative_2019} library and was optimized using the implementation of the global \textit{PSO} of the pyswarms library \cite{pyswarmsJOSS2018} for the pre-processed \textit{BSDS500} dataset. For cross-validation in k-folds, the implementation of the library \textit{scikit-learn} \cite{scikit-learn} was used.

For all the experiments defined in this research, the edge maps of the pre-processed dataset were extracted for the threshold of probability. The probability threshold will be referred to in the course of the text as $p$.

The model will be evaluated using the following case studies:

\begin{enumerate}
    \item \label{r_k_fold} find the $\tau$, the $\Delta$ and the optimal rule of the automaton for $r \in \{1,2\}$ combined with cross-validation in k-folds where $k = 10$;
    \item \label{r1_general} find the $\tau$, the $\Delta$ and the rule optimized for $r = 1$ using as a training set all the images and save the final population and the best rule;
    \begin{enumerate}
        \item test the optimal rule found from step \ref{r1_general} in the set of all images;
        \item test the optimal rule found of step \ref{r1_general} in the set of each category separately;
    \end{enumerate}
    \item \label{r2_general} find the $\tau$, the $\Delta$ and the rule optimized for $r = 2$ using all images as a training set and save the final population and the best rule.
    \begin{enumerate}
        \item test the optimal rule found from step \ref{r2_general} in the set of all images;
        \item test the optimal rule found of step \ref{r2_general} in the set of each category separately;
    \end{enumerate}
    \item \label{r1_tf} using as initial population the final population of step \ref{r1_general}, search separately for each category the $\tau$, the $\Delta$ and the automaton rule of the same space and evaluate for all categories;
    \item \label{r2_tf} using as initial population the final population of step \ref{r2_general}, search separately for each category the $\tau$, the $\Delta$ and the automaton rule of the same space and evaluate for all categories;
    \item \label{r_individual} search each category (separately, and without using the rules previously found by the other experiments) for a rule such that the radius $r \in \{1,2\}$ and test the rule found in each category for all categories;
\end{enumerate}

The purpose of this process is to use several samples from the same dataset to evaluate the robustness of the edge detection task. After the execution of the tests, the results of the selected evaluation metrics will be compared between the detector of the proposed model optimized for each value of $r$ and the Canny edge detector implemented by the \textit{skimage} library using the base parameters of its implementation, where \(\sigma = 1\), the lower and upper thresholds are equivalent to 10\% and 20\% of the maximum value of the image type, which for the search is \textit{int8} and is equivalent to 255, the maximum value of grayscale images.

\subsubsection{Metrics}
To evaluate the fitness of the particles when applied to a rule, the Dice Similarity Score \textit{(DSC)} is calculated from the annotated edge map and the edge map extracted by the current rule. The metric can be described by the formula:
\begin{equation}
    DSC = (2 \cdot TP)/(2 \cdot TP + FP + FN),
\end{equation}
For the edge detection problem, the positive class is defined as an edge and the negative as non-edge. The terms of the previous formula are described as:
\begin{itemize}
    \item TP (True Positive) represents that the detector correctly classified the image pixel as a edge;
    \item FP (False Positive) represents that the detector incorrectly classified the image pixel as a edge;
    \item FN (False Negative) represents that the detector did not classify the pixel of the image as a edge and in the annotation of the image the pixel is classified as a edge;
\end{itemize}

The quality of the detected edges is evaluated by the Peak signal-to-noise ratio metric that signals the ratio of noise to the amount of information in the original image \cite{hore2013there}.
The formula is defined by:

\begin{equation}
    PSNR(f, g) = 10 \log_{10} \left(\frac{R^{2}}{MSE}\right),
\end{equation}

In which (MSE) is described by:
\begin{equation}
    MSE(f, g) = \frac{1}{MN}\sum_{i=1}^{M}\sum_{j=1}^{N}(f_{ij} - g_{ij})^2,
\end{equation}

In the formula described, $f$ corresponds to the annotated edge map and $g$ to the detected edge map, both with dimensions \(M\times N\). $R$ corresponds to the maximum variation value of the image. For this study that works only with gray scale images, $R$ will be considered as $255$.
The calculation of the Mean Squared Error (MSE) compares the values of the annotated edge map with the variation of the detected edge map, and the resulting value represents the amount of noise in the variation of the edge map. High metric values indicate that the error rate is low and there is little noise between the image information.
Despite being a widely used metric in the area of compression and retrieval of information in images, it is also applied to evaluate edge detectors \cite{padmavathi2009performance, poobathy2014edge,dumitru2021transfer} and image segmentation \cite{xess2014analysis}.

The Structural Similarity (SSIM) metric evaluates the similarity between the detector result and the annotated edge map \cite{wang2004image}. The value of the metric is defined in the range $[-1, 1]$, with values equal to 1 indicating that both annotated and detected edge maps are similar, 0 representing that there is no similarity, and -1 indicating the existence of a negative correlation in similarity.  
The metric formula is described by:
\begin{equation}
    SSIM(x, y) = l(x, y) \cdot c(x, y) \cdot s(x, y),
\end{equation}
where $l$ corresponds to the comparison of luminance, $c$ to the contrast and $s$ the structure of the images $x$ and $y$.

\section{Discussion}

\subsection{k-fold Cross-validation result analysis}
\begin{figure}[!htbp]
    \centering
    \subfigure[\textit{SSIM}\label{fig:pso-caxcanny-fold-ssim-0.02-avg}]{\includegraphics[width=\textwidth]{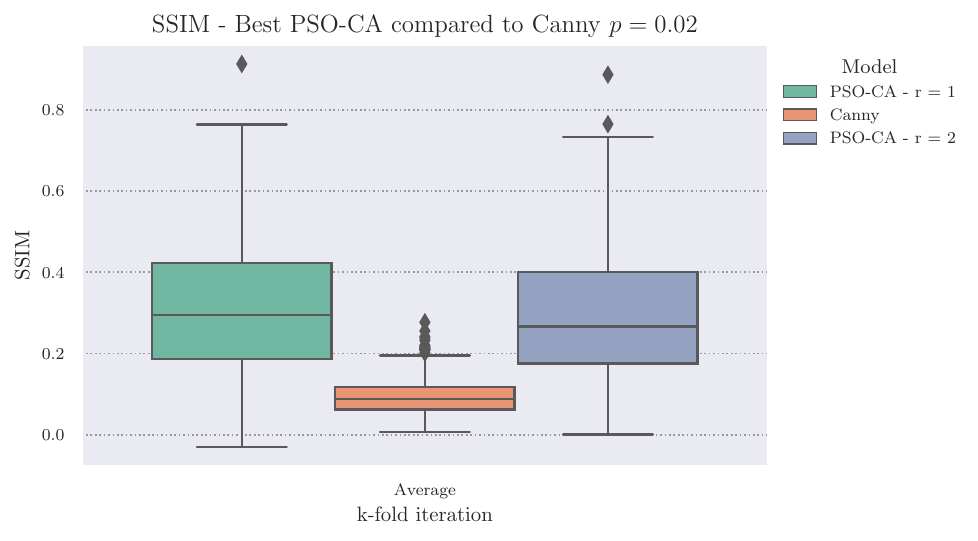}}
    \subfigure[\textit{PSNR}\label{fig:pso-caxcanny-fold-psnr-0.02-avg}]{
        \includegraphics[width=\textwidth]{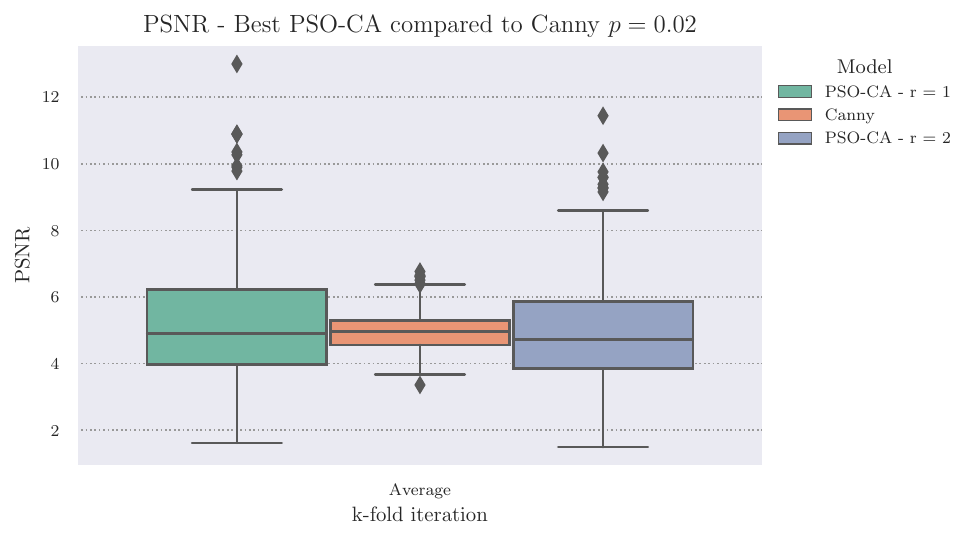}}
    \caption{Comparison of the values of the \textit{SSIM} and \textit{PSNR} metric between the best model of the iteration of the k-folders and the Canny detector for each iteration of the k-folds.}
    \label{fig:pso-caxcanny-fold-0.02-avg}
\end{figure}

The Figure \ref{fig:pso-caxcanny-fold-0.02-avg} and Table \ref{tab1:resultados_0.02} present the average results of the k-fold cross-validation for the case study \ref{r_k_fold}. It’s evident from the results that there’s significant variation and the presence of some outliers for both chosen metrics, as per each iteration.

Despite the variations in the two metrics, a constant was the model with $r = 1$ presenting better average values when compared to the model with $r = 2$ (figure \ref{fig:pso-caxcanny-fold-0.02-avg}).

\begin{table}[!hbp]
\centering
\caption{Results of the average value of all iterations of k-fold for the \textit{PSO-CA} model and the Canny model}
\label{tab1:resultados_0.02}
\newcolumntype{C}{>{\centering\arraybackslash}X}
\begin{tabular}{ccccccc}
\toprule
Model & $r$ & $p$ & \multicolumn{2}{c}{\textit{PSNR}} & \multicolumn{2}{c}{SSIM} \\
\midrule
\textit{PSO-CA} & 1 & 0,02 & 5,145 & $\pm$ 1,707 & 0,312 & $\pm$ 0,157 \\
\textit{PSO-CA} & 2 & 0,02 & 4,897 & $\pm$ 1,554 & 0,292 & $\pm$ 0,150 \\
 Canny & - & 0,02 & 4,948 & $\pm$ 0,561 & 0,093 & $\pm$ 0,044 \\
\bottomrule
\end{tabular}
\end{table}

\begin{table}[!htpb]
\caption{Results of the objective function of the optimization step and the values of the parameters found of all iterations of the k-folds for the \textit{PSO-CA} model and map with probability threshold equal to 0.02.}
\label{tab3:resultados_otimizacao_pso_ca_k_fold_0.02}
\newcolumntype{C}{>{\centering\arraybackslash}X}
\begin{tabularx}{\textwidth}{*{8}{C}}
\toprule
Model & $r$ & $p$ & k-fold iteration & $\Delta$ & $\tau$ & $z$ & \textit{DSC} \\
\midrule
\multirow[t]{20}{*}{\textit{PSO-CA}} & \multirow[t]{10}{*}{1} & \multirow[t]{10}{*}{0,02} & 0 & 20 & 0,744077 & 350 & 0,517 \\
\cline{4-8} \cline{5-8} \cline{6-8}
 &  &  & 1 & 127 & 0,338653 & 447 & 0,515 \\
\cline{4-8} \cline{5-8} \cline{6-8}
 &  &  & 2 & 126 & 0,267417 & 158 & 0,507 \\
\cline{4-8} \cline{5-8} \cline{6-8}
 &  &  & 3 & 125 & 0,325631 & 351 & 0,507 \\
\cline{4-8} \cline{5-8} \cline{6-8}
 &  &  & 4 & 146 & 0,280605 & 414 & 0,519 \\
\cline{4-8} \cline{5-8} \cline{6-8}
 &  &  & 5 & 120 & 0,353144 & 446 & 0,513 \\
\cline{4-8} \cline{5-8} \cline{6-8}
 &  &  & 6 & 121 & 0,330819 & 350 & 0,510 \\
\cline{4-8} \cline{5-8} \cline{6-8}
 &  &  & 7 & 234 & 0,206608 & 350 & 0,510 \\
\cline{4-8} \cline{5-8} \cline{6-8}
 &  &  & 8 & 231 & 0,214776 & 447 & 0,517 \\
\cline{4-8} \cline{5-8} \cline{6-8}
 &  &  & 9 & 253 & 0,206034 & 446 & 0,510 \\
\cline{2-8} \cline{3-8} \cline{4-8} \cline{5-8} \cline{6-8}
 & \multirow[t]{10}{*}{2} & \multirow[t]{10}{*}{0,02} & 0 & 198 & 0,379419 & 566719 & 0,510 \\
\cline{4-8} \cline{5-8} \cline{6-8}
 &  &  & 1 & 65 & 0,751934 & 4390783 & 0,505 \\
\cline{4-8} \cline{5-8} \cline{6-8}
 &  &  & 2 & 86 & 0,590864 & 4246846 & 0,499 \\
\cline{4-8} \cline{5-8} \cline{6-8}
 &  &  & 3 & 129 & 0,535537 & 8435166 & 0,498 \\
\cline{4-8} \cline{5-8} \cline{6-8}
 &  &  & 4 & 80 & 0,543195 & 36317 & 0,514 \\
\cline{4-8} \cline{5-8} \cline{6-8}
 &  &  & 5 & 197 & 0,432928 & 12642622 & 0,501 \\
\cline{4-8} \cline{5-8} \cline{6-8}
 &  &  & 6 & 108 & 0,490182 & 535990 & 0,500 \\
\cline{4-8} \cline{5-8} \cline{6-8}
 &  &  & 7 & 51 & 0,677149 & 44508 & 0,504 \\
\cline{4-8} \cline{5-8} \cline{6-8}
 &  &  & 8 & 158 & 0,483424 & 8447358 & 0,508 \\
\cline{4-8} \cline{5-8} \cline{6-8}
 &  &  & 9 & 217 & 0,398464 & 556538 & 0,501 \\
\cline{1-8} \cline{2-8} \cline{3-8} \cline{4-8} \cline{5-8} \cline{6-8}
\bottomrule
\end{tabularx}

\end{table}

When analyzing the variation of the fitness function results for the particle at the end of each training iteration (Table \ref{tab3:resultados_otimizacao_pso_ca_k_fold_0.02}) it becomes evident that despite the differences in the values of \textit{DSC}, both values of $r$ exhibited consistent fluctuations across each fold.

The table \ref{tab3:resultados_otimizacao_pso_ca_k_fold_0.02} underscores the adaptability of the parameters in finding a rule. The folds of iterations 0, 6, and 7 converged on the same rule (350), while the variation in the \(\Delta\) and \(\tau\) parameters enabled the model to adapt to the characteristics of images of each fold.

The proposed model had a greater variation for the \textit{PSNR} metric when compared to Canny. However, it still showed better results for most scenarios. For the \textit{SSIM} metric, the proposed model was able to present considerably better results in all iterations of all study cases.

\subsection{Cross-evaluation in the categories}
\begin{figure}[!htbp]  
\centering 
        \includegraphics[width=\textwidth]{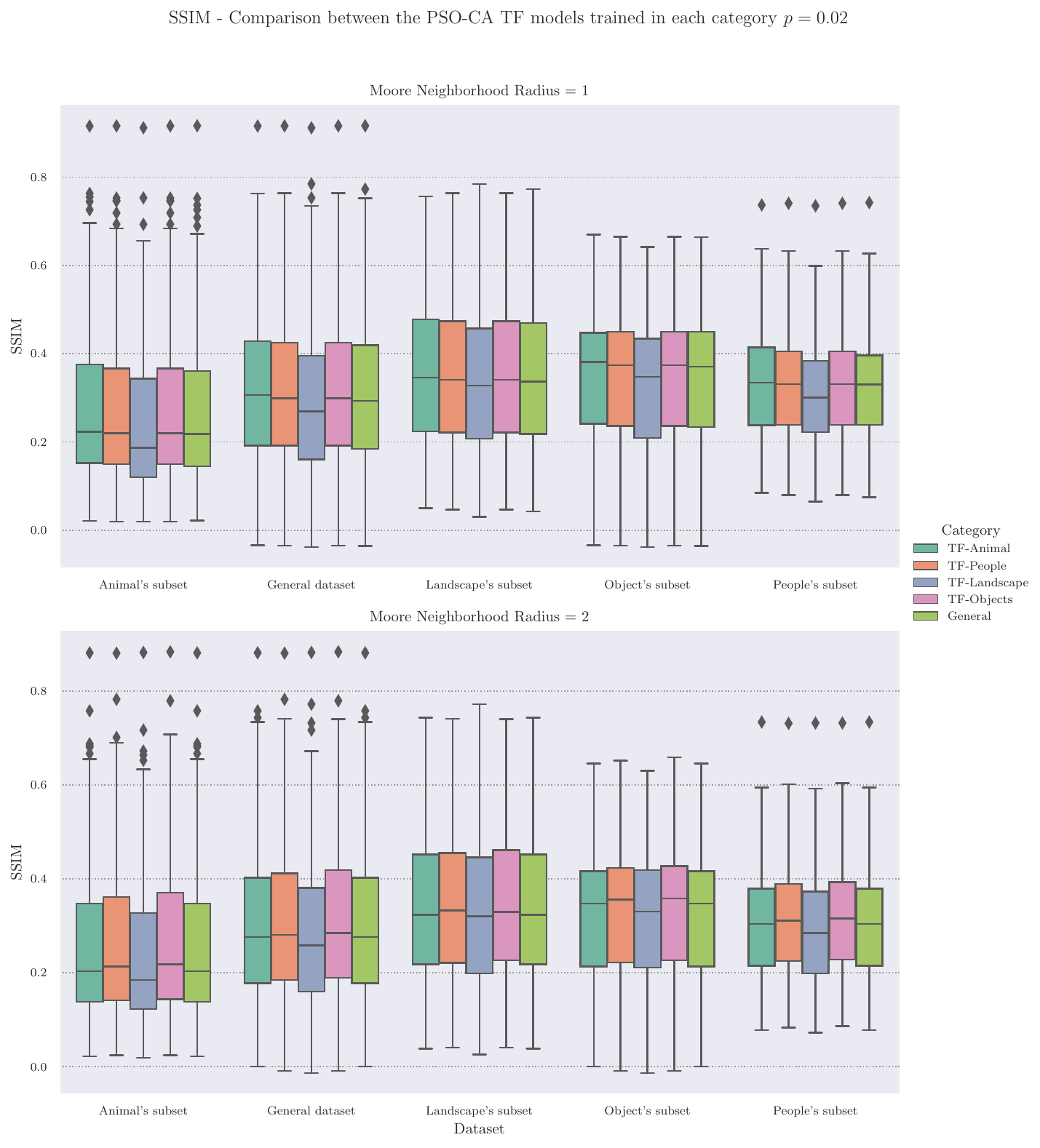}
    \caption{Comparison of the values of the \textit{SSIM} metric for the study cases: \ref{r1_general} \ref{r2_general}, \ref{r1_tf} and \ref{r2_tf}. For each value of $r$ and each category of the dataset, the general models are compared with the specialized models.}
    \label{fig:pso-ca-r1xr2-categories-ssim-0.02}
\end{figure}

\begin{figure}[!htbp]
    \centering 
    \includegraphics[width=\textwidth]{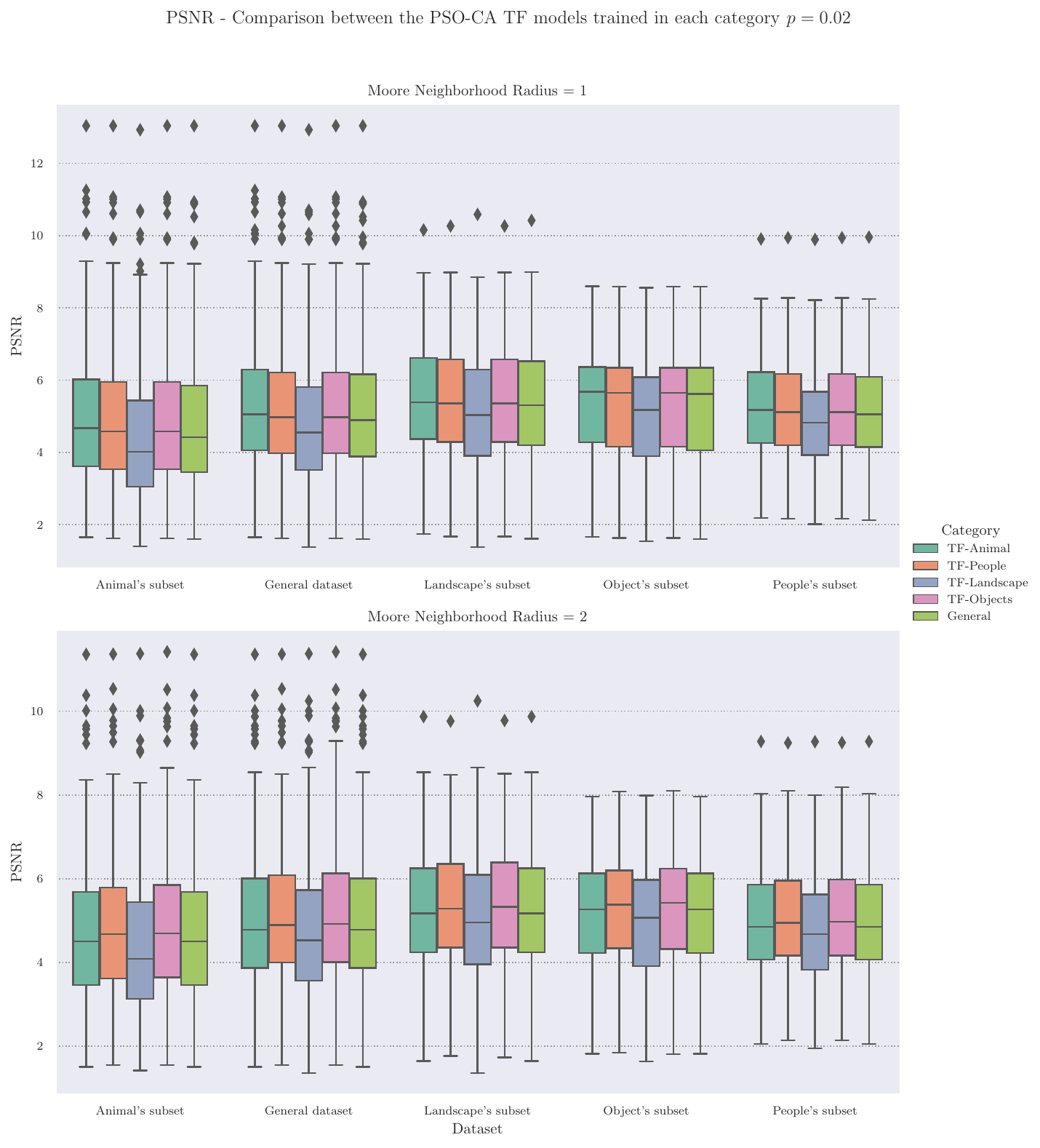}
    \caption{Comparison of the values of the \textit{PSNR} metric for the case studies: \ref{r1_general} \ref{r2_general}, \ref{r1_tf} and \ref{r2_tf}. For each value of $r$ and each category of the dataset, the general models are compared with the specialized models.}
    \label{fig:pso-ca-r1xr2-categories-psnr-0.02}
\end{figure}

The case studies \ref{r1_tf} and \ref{r2_tf} delve into the exploration of the rule and the adaptation of the model to the categories present in the pre-processed dataset. The Figures \ref{fig:pso-ca-r1xr2-categories-ssim-0.02} and \ref{fig:pso-ca-r1xr2-categories-psnr-0.02} and Table \ref{tab9:resultados_comparacao_pso_ca_0.02} present the average result for both metrics, \textit{SSIM} and \textit{PSNR}, when comparing the general model, the model trained using the entire dataset, and the specialized model for each category for each subset of the dataset. Additionally, Table \ref{tab5:resultados_canny_categories_0.02} presents the results obtained by Canny.

For the case studies \ref{r1_tf} and \ref{r2_tf}, the results of the application of Transfer Learning were represented by the prefix \textit{TF} in the name of each category.

When comparing the proposed \textit{PSO-CA} models with Canny (Table \ref{tab5:resultados_canny_categories_0.02}), the proposed models were able to present better results for both \textit{PSNR} and \textit{SSIM}, indicating that they were able to adapt to individual categories of the images.

The results obtained also indicate that the general category model and the specialized model of each category were able to find different rules (Table \ref{tab10:resultados_comparacao_pso_ca_dsc_0.02}) and in some cases, the new rule adapted better to the category as for the models trained on animals and objects. When comparing the values of \textit{DSC} of the rule found in the optimization stage of the general model with the specialized models, it is noticed that the model specialized in animals presented values of \textit{DSC} below the general model despite having good results for both evaluation metrics.

\begin{table}[!htpb]
\centering
\fontsize{5.9pt}{6.25pt}\selectfont
\caption{Comparison between the average results of each evaluation metric for each PSO-CA experiment}
\label{tab9:resultados_comparacao_pso_ca_0.02}
\newcolumntype{C}{>{\centering\arraybackslash}X}
\begin{tabularx}{\textwidth}{*{8}{C}}
\toprule
Model & $r$ & $p$ & Experiment & \multicolumn{2}{c}{\textit{PSNR}} & \multicolumn{2}{c}{\textit{SSIM}}  \\
\midrule
\multirow[t]{20}{*}{PSO-CA} & \multirow[t]{10}{*}{1} & \multirow[t]{10}{*}{0,02} & TF-Animals Model& 5,245 & $\pm$ 1,706 & 0,319 & $\pm$ 0,157 \\
\cline{4-8}
 &  &  & TF-Objects Model & 5,176 & $\pm$ 1,715 & 0,315 & $\pm$ 0,157 \\
\cline{4-8}
 &  &  & TF-Landscapes Model & 4,789 & $\pm$ 1,720 & 0,289 & $\pm$ 0,156 \\
\cline{4-8}
 &  &  & TF-People Model & 5,176 & $\pm$ 1,715 & 0,315 & $\pm$ 0,157 \\
\cline{4-8}
 &  &  & General model trained on the entire dataset & 5,105 & $\pm$ 1,723 & 0,311 & $\pm$ 0,157 \\
\cline{4-8}
 &  &  & Model trained only on Animals & 5,278 & $\pm$ 1,702 & 0,321 & $\pm$ 0,157 \\
\cline{4-8}
 &  &  & Model trained only on Objects & 5,094 & $\pm$ 1,697 & 0,305 & $\pm$ 0,157 \\
\cline{4-8}
 &  &  & Model trained only on Landscapes & 4,789 & $\pm$ 1,720 & 0,289 & $\pm$ 0,156 \\
\cline{4-8}
 &  &  & Model trained only on People & 5,217 & $\pm$ 1,688 & 0,315 & $\pm$ 0,156 \\
\cline{4-8}
 &  &  & Average 10-fold & 5,145 & $\pm$ 1,707 & 0,312 & $\pm$ 0,157 \\
\cline{2-8} \cline{3-8}
 & \multirow[t]{10}{*}{2} & \multirow[t]{10}{*}{0,02} & TF-Animals Model & 4,968 & $\pm$ 1,569 & 0,297 & $\pm$ 0,152 \\
 \cline{4-8}
 &  &  & TF-Objects Model & 5,101 & $\pm$ 1,572 & 0,306 & $\pm$ 0,153 \\
 \cline{4-8}
 &  &  & TF-Landscapes Model & 4,710 & $\pm$ 1,596 & 0,281 & $\pm$ 0,152 \\
 \cline{4-8}
 &  &  & TF-People Model & 5,081 & $\pm$ 1,555 & 0,303 & $\pm$ 0,152 \\
 \cline{4-8}
 &  &  & General model trained on the entire dataset & 4,968 & $\pm$ 1,569 & 0,297 & $\pm$ 0,152 \\
 \cline{4-8}
 &  &  & Model trained only on Animals & 4,838 & $\pm$ 1,560 & 0,288 & $\pm$ 0,151 \\
 \cline{4-8}
 &  &  & Model trained only on Objects & 5,049 & $\pm$ 1,530 & 0,297 & $\pm$ 0,150 \\
 \cline{4-8}
 &  &  & Model trained only on Landscapes & 4,566 & $\pm$ 1,523 & 0,267 & $\pm$ 0,148 \\
 \cline{4-8}
 &  &  & Model trained only on People & 4,975 & $\pm$ 1,541 & 0,297 & $\pm$ 0,151 \\
 \cline{4-8}
 &  &  & Average 10-fold & 4,897 & $\pm$ 1,554 & 0,292 & $\pm$ 0,150 \\
\cline{1-8} \cline{2-8} \cline{3-8}
\bottomrule
\end{tabularx}
\end{table}

\begin{table}[!htpb]
\centering
\caption{Average results for both evaluation metrics for Canny for each category of the dataset}
\label{tab5:resultados_canny_categories_0.02}
\newcolumntype{C}{>{\centering\arraybackslash}X}
\begin{tabularx}{\textwidth}{*{7}{C}}
\toprule
Model & Dataset & $p$ &  \multicolumn{2}{c}{\textit{PSNR}} & \multicolumn{2}{c}{\textit{SSIM}}  \\
\midrule
\multirow[t]{5}{*}{Canny} & Animals & 0,02 & 4,922 & $\pm$ 0,610 & 0,077 & $\pm$ 0,041 \\
\cline{2-7}
 & General & 0,02 & 4,948 & $\pm$ 0,561 & 0,093 & $\pm$ 0,044 \\
\cline{2-7}
 & Landscapes & 0,02 & 5,035 & $\pm$ 0,587 & 0,094 & $\pm$ 0,031 \\
\cline{2-7}
 & Objects & 0,02 & 4,993 & $\pm$ 0,478 & 0,107 & $\pm$ 0,050 \\
\cline{2-7}
 & People & 0,02 & 4,907 & $\pm$ 0,494 & 0,108 & $\pm$ 0,044 \\
\cline{1-7} \cline{2-7}
\bottomrule
\end{tabularx}
\end{table}

\begin{table}[!htpb]
\fontsize{4.0pt}{4.75pt}\selectfont
\centering
\caption{Average results of both metrics for each category of the dataset for the experiments \ref{r1_general}, \ref{r2_general}, \ref{r1_tf}, \ref{r2_tf}, and \ref{r_individual} }
\label{tab11:resultados_comparacao_pso_ca_categories_0.02}
\newcolumntype{C}{>{\centering\arraybackslash}X}
\begin{tabularx}{\textwidth}{*{9}{C}}
\toprule
Model & $r$ & $p$ & Dataset & Training Dataset & \multicolumn{2}{c}{\textit{PSNR}} & \multicolumn{2}{c}{\textit{SSIM}} \\
\midrule

\multirow[t]{90}{*}{PSO-CA} & \multirow[t]{45}{*}{1} & \multirow[t]{45}{*}{0,02} & \multirow[t]{9}{*}{Animals} & Animals & 5,116 & $\pm$ 2,028 & 0,282 & $\pm$ 0,175 \\
\cline{5-9}
 &  &  &  & General & 4,903 & $\pm$ 2,038 & 0,270 & $\pm$ 0,173 \\
\cline{5-9}
 &  &  &  & Objects & 4,865 & $\pm$ 1,996 & 0,260 & $\pm$ 0,170 \\
\cline{5-9}
 &  &  &  & Landscapes & 4,514 & $\pm$ 1,998 & 0,241 & $\pm$ 0,166 \\
\cline{5-9}
 &  &  &  & People & 5,050 & $\pm$ 2,007 & 0,274 & $\pm$ 0,173 \\
\cline{5-9}
 &  &  &  & TF-Animals & 5,075 & $\pm$ 2,031 & 0,279 & $\pm$ 0,175 \\
\cline{5-9}
 &  &  &  & TF-Objects & 4,990 & $\pm$ 2,035 & 0,274 & $\pm$ 0,174 \\
\cline{5-9}
 &  &  &  & TF-Landscapes & 4,514 & $\pm$ 1,998 & 0,241 & $\pm$ 0,166 \\
\cline{5-9}
 &  &  &  & TF-People & 4,990 & $\pm$ 2,035 & 0,274 & $\pm$ 0,174 \\
\cline{4-9} \cline{5-9}
 &  &  & \multirow[t]{9}{*}{General} & Animals & 5,278 & $\pm$ 1,702 & 0,321 & $\pm$ 0,157 \\
\cline{5-9}
 &  &  &  & General & 5,105 & $\pm$ 1,723 & 0,311 & $\pm$ 0,157 \\
\cline{5-9}
 &  &  &  & Objects & 5,094 & $\pm$ 1,697 & 0,305 & $\pm$ 0,157 \\
\cline{5-9}
 &  &  &  & Landscapes & 4,789 & $\pm$ 1,720 & 0,289 & $\pm$ 0,156 \\
\cline{5-9}
 &  &  &  & People & 5,217 & $\pm$ 1,688 & 0,315 & $\pm$ 0,156 \\
\cline{5-9}
 &  &  &  & TF-Animals & 5,245 & $\pm$ 1,706 & 0,319 & $\pm$ 0,157 \\
\cline{5-9}
 &  &  &  & TF-Objects & 5,176 & $\pm$ 1,715 & 0,315 & $\pm$ 0,157 \\
\cline{5-9}
 &  &  &  & TF-Landscapes & 4,789 & $\pm$ 1,720 & 0,289 & $\pm$ 0,156 \\
\cline{5-9}
 &  &  &  & TF-People & 5,176 & $\pm$ 1,715 & 0,315 & $\pm$ 0,157 \\
\cline{4-9} \cline{5-9}
 &  &  & \multirow[t]{9}{*}{Objects} & Animals & 5,463 & $\pm$ 1,564 & 0,355 & $\pm$ 0,148 \\
\cline{5-9}
 &  &  &  & General & 5,316 & $\pm$ 1,600 & 0,348 & $\pm$ 0,151 \\
\cline{5-9}
 &  &  &  & Objects & 5,330 & $\pm$ 1,586 & 0,345 & $\pm$ 0,150 \\
\cline{5-9}
 &  &  &  & Landscapes & 5,044 & $\pm$ 1,609 & 0,328 & $\pm$ 0,152 \\
\cline{5-9}
 &  &  &  & People & 5,406 & $\pm$ 1,554 & 0,350 & $\pm$ 0,149 \\
\cline{5-9}
 &  &  &  & TF-Animals & 5,434 & $\pm$ 1,571 & 0,353 & $\pm$ 0,149 \\
\cline{5-9}
 &  &  &  & TF-Objects & 5,376 & $\pm$ 1,587 & 0,351 & $\pm$ 0,150 \\
\cline{5-9}
 &  &  &  & TF-Landscapes & 5,044 & $\pm$ 1,609 & 0,328 & $\pm$ 0,152 \\
\cline{5-9}
 &  &  &  & TF-People & 5,376 & $\pm$ 1,587 & 0,351 & $\pm$ 0,150 \\
\cline{4-9} \cline{5-9}
 &  &  & \multirow[t]{9}{*}{Landscapes} & Animals & 5,489 & $\pm$ 1,610 & 0,354 & $\pm$ 0,158 \\
\cline{5-9}
 &  &  &  & General & 5,350 & $\pm$ 1,660 & 0,348 & $\pm$ 0,161 \\
\cline{5-9}
 &  &  &  & Objects & 5,356 & $\pm$ 1,641 & 0,345 & $\pm$ 0,160 \\
\cline{5-9}
 &  &  &  & Landscapes & 5,106 & $\pm$ 1,713 & 0,333 & $\pm$ 0,163 \\
\cline{5-9}
 &  &  &  & People & 5,443 & $\pm$ 1,603 & 0,349 & $\pm$ 0,158 \\
\cline{5-9}
 &  &  &  & TF-Animals & 5,463 & $\pm$ 1,620 & 0,353 & $\pm$ 0,159 \\
\cline{5-9}
 &  &  &  & TF-Objects & 5,407 & $\pm$ 1,638 & 0,350 & $\pm$ 0,160 \\
\cline{5-9}
 &  &  &  & TF-Landscapes & 5,106 & $\pm$ 1,713 & 0,333 & $\pm$ 0,163 \\
\cline{5-9}
 &  &  &  & TF-People & 5,407 & $\pm$ 1,638 & 0,350 & $\pm$ 0,160 \\
\cline{4-9} \cline{5-9}
 &  &  & \multirow[t]{9}{*}{People} & Animals & 5,283 & $\pm$ 1,292 & 0,338 & $\pm$ 0,122 \\
\cline{5-9}
 &  &  &  & General & 5,127 & $\pm$ 1,293 & 0,328 & $\pm$ 0,121 \\
\cline{5-9}
 &  &  &  & Objects & 5,130 & $\pm$ 1,275 & 0,324 & $\pm$ 0,122 \\
\cline{5-9}
 &  &  &  & Landscapes & 4,844 & $\pm$ 1,276 & 0,307 & $\pm$ 0,121 \\
\cline{5-9}
 &  &  &  & People & 5,215 & $\pm$ 1,279 & 0,331 & $\pm$ 0,121 \\
\cline{5-9}
 &  &  &  & TF-Animals & 5,254 & $\pm$ 1,291 & 0,336 & $\pm$ 0,121 \\
\cline{5-9}
 &  &  &  & TF-Objects & 5,193 & $\pm$ 1,294 & 0,333 & $\pm$ 0,121 \\
\cline{5-9}
 &  &  &  & TF-Landscapes & 4,844 & $\pm$ 1,276 & 0,307 & $\pm$ 0,121 \\
\cline{5-9}
 &  &  &  & TF-People & 5,193 & $\pm$ 1,294 & 0,333 & $\pm$ 0,121 \\
\cline{2-9} \cline{3-9} \cline{4-9} \cline{5-9}
 & \multirow[t]{45}{*}{2} & \multirow[t]{45}{*}{0,02} & \multirow[t]{9}{*}{Animals} & Animals & 4,685 & $\pm$ 1,818 & 0,254 & $\pm$ 0,168 \\
\cline{5-9}
 &  &  &  & General & 4,818 & $\pm$ 1,830 & 0,261 & $\pm$ 0,168 \\
\cline{5-9}
 &  &  &  & Objects & 4,916 & $\pm$ 1,789 & 0,262 & $\pm$ 0,167 \\
\cline{5-9}
 &  &  &  & Landscapes & 4,374 & $\pm$ 1,775 & 0,232 & $\pm$ 0,163 \\
\cline{5-9}
 &  &  &  & People & 4,843 & $\pm$ 1,802 & 0,264 & $\pm$ 0,168 \\
\cline{5-9}
 &  &  &  & TF-Animals & 4,818 & $\pm$ 1,830 & 0,261 & $\pm$ 0,168 \\
\cline{5-9}
 &  &  &  & TF-Objects & 4,963 & $\pm$ 1,841 & 0,270 & $\pm$ 0,170 \\
\cline{5-9}
 &  &  &  & TF-Landscapes & 4,494 & $\pm$ 1,842 & 0,241 & $\pm$ 0,164 \\
\cline{5-9}
 &  &  &  & TF-People & 4,942 & $\pm$ 1,819 & 0,267 & $\pm$ 0,169 \\
\cline{4-9} \cline{5-9}
 &  &  & \multirow[t]{9}{*}{General} & Animals & 4,838 & $\pm$ 1,560 & 0,288 & $\pm$ 0,151 \\
\cline{5-9}
 &  &  &  & General & 4,968 & $\pm$ 1,569 & 0,297 & $\pm$ 0,152 \\
\cline{5-9}
 &  &  &  & Objects & 5,049 & $\pm$ 1,530 & 0,297 & $\pm$ 0,150 \\
\cline{5-9}
 &  &  &  & Landscapes & 4,566 & $\pm$ 1,523 & 0,267 & $\pm$ 0,148 \\
\cline{5-9}
 &  &  &  & People & 4,975 & $\pm$ 1,541 & 0,297 & $\pm$ 0,151 \\
\cline{5-9}
 &  &  &  & TF-Animals & 4,968 & $\pm$ 1,569 & 0,297 & $\pm$ 0,152 \\
\cline{5-9}
 &  &  &  & TF-Objects & 5,101 & $\pm$ 1,572 & 0,306 & $\pm$ 0,153 \\
\cline{5-9}
 &  &  &  & TF-Landscapes & 4,710 & $\pm$ 1,596 & 0,281 & $\pm$ 0,152 \\
\cline{5-9}
 &  &  &  & TF-People & 5,081 & $\pm$ 1,555 & 0,303 & $\pm$ 0,152 \\
\cline{4-9} \cline{5-9}
 &  &  & \multirow[t]{9}{*}{Objects} & Animals & 4,994 & $\pm$ 1,419 & 0,318 & $\pm$ 0,139 \\
\cline{5-9}
 &  &  &  & General & 5,130 & $\pm$ 1,439 & 0,328 & $\pm$ 0,141 \\
\cline{5-9}
 &  &  &  & Objects & 5,214 & $\pm$ 1,399 & 0,329 & $\pm$ 0,140 \\
\cline{5-9}
 &  &  &  & Landscapes & 4,740 & $\pm$ 1,384 & 0,299 & $\pm$ 0,138 \\
\cline{5-9}
 &  &  &  & People & 5,105 & $\pm$ 1,387 & 0,325 & $\pm$ 0,138 \\
\cline{5-9}
 &  &  &  & TF-Animals & 5,130 & $\pm$ 1,439 & 0,328 & $\pm$ 0,141 \\
\cline{5-9}
 &  &  &  & TF-Objects & 5,260 & $\pm$ 1,447 & 0,338 & $\pm$ 0,143 \\
\cline{5-9}
 &  &  &  & TF-Landscapes & 4,913 & $\pm$ 1,480 & 0,315 & $\pm$ 0,144 \\
\cline{5-9}
 &  &  &  & TF-People & 5,232 & $\pm$ 1,420 & 0,334 & $\pm$ 0,140 \\
\cline{4-9} \cline{5-9}
 &  &  & \multirow[t]{9}{*}{Landscapes} & Animals & 5,101 & $\pm$ 1,529 & 0,327 & $\pm$ 0,156 \\
\cline{5-9}
 &  &  &  & General & 5,216 & $\pm$ 1,532 & 0,334 & $\pm$ 0,156 \\
\cline{5-9}
 &  &  &  & Objects & 5,280 & $\pm$ 1,482 & 0,333 & $\pm$ 0,154 \\
\cline{5-9}
 &  &  &  & Landscapes & 4,828 & $\pm$ 1,500 & 0,305 & $\pm$ 0,154 \\
\cline{5-9}
 &  &  &  & People & 5,198 & $\pm$ 1,501 & 0,331 & $\pm$ 0,155 \\
\cline{5-9}
 &  &  &  & TF-Animals & 5,216 & $\pm$ 1,532 & 0,334 & $\pm$ 0,156 \\
\cline{5-9}
 &  &  &  & TF-Objects & 5,332 & $\pm$ 1,518 & 0,342 & $\pm$ 0,156 \\
\cline{5-9}
 &  &  &  & TF-Landscapes & 5,014 & $\pm$ 1,600 & 0,324 & $\pm$ 0,160 \\
\cline{5-9}
 &  &  &  & TF-People & 5,313 & $\pm$ 1,505 & 0,339 & $\pm$ 0,156 \\
\cline{4-9} \cline{5-9}
 &  &  & \multirow[t]{9}{*}{People} & Animals & 4,807 & $\pm$ 1,217 & 0,296 & $\pm$ 0,118 \\
\cline{5-9}
 &  &  &  & General & 4,941 & $\pm$ 1,226 & 0,307 & $\pm$ 0,119 \\
\cline{5-9}
 &  &  &  & Objects & 5,010 & $\pm$ 1,199 & 0,307 & $\pm$ 0,116 \\
\cline{5-9}
 &  &  &  & Landscapes & 4,577 & $\pm$ 1,175 & 0,276 & $\pm$ 0,114 \\
\cline{5-9}
 &  &  &  & People & 4,951 & $\pm$ 1,213 & 0,307 & $\pm$ 0,118 \\
\cline{5-9}
 &  &  &  & TF-Animals & 4,941 & $\pm$ 1,226 & 0,307 & $\pm$ 0,119 \\
\cline{5-9}
 &  &  &  & TF-Objects & 5,069 & $\pm$ 1,227 & 0,317 & $\pm$ 0,119 \\
\cline{5-9}
 &  &  &  & TF-Landscapes & 4,715 & $\pm$ 1,219 & 0,292 & $\pm$ 0,118 \\
\cline{5-9}
 &  &  &  & TF-People & 5,052 & $\pm$ 1,218 & 0,314 & $\pm$ 0,119 \\
\cline{1-9} \cline{2-9} \cline{3-9} \cline{4-9} \cline{5-9}
\bottomrule
\end{tabularx}
\end{table}

    \begin{table}
    \centering
    \fontsize{5.5pt}{7.5pt}\selectfont
    \caption{Comparison between the results found for the fitness function at the end of the optimization step for experiments \ref{r1_general}, \ref{r2_general}, \ref{r1_tf}, \ref{r2_tf}, and \ref{r_individual}}
    \label{tab10:resultados_comparacao_pso_ca_dsc_0.02}
    \begin{tabular}{lllllllrr}
    \toprule
    Model & $r$ & $p$ & Experiment & $\Delta$  & $\tau$ & $z$ & \multicolumn{2}{c}{DSC} \\
    \midrule
    \multirow[t]{38}{*}{PSO-CA} & \multirow[t]{19}{*}{1} & \multirow[t]{19}{*}{0,02} & General TF model trained in the Animal's subset & 89 & 0,431699 & 446 & 0,467 & 0,000 \\
    \cline{4-9} \cline{5-9} \cline{6-9}
     &  &  & General TF model trained in the Landscape's subset & 74 & 0,398369 & 415 & 0,532 & 0,000 \\
    \cline{4-9} \cline{5-9} \cline{6-9}
     &  &  & General TF model trained in the Object's subset & 82 & 0,445122 & 447 & 0,543 & 0,000 \\
    \cline{4-9} \cline{5-9} \cline{6-9}
     &  &  & General TF model trained in the People's subset & 91 & 0,419048 & 446 & 0,552 & 0,000 \\
    \cline{4-9} \cline{5-9} \cline{6-9}
     &  &  & General model trained in General dataset & 102 & 0,382719 & 447 & 0,513 & 0,000 \\
    \cline{4-9} \cline{5-9} \cline{6-9}
     &  &  & Model trained only in the Animal's subset & 75 & 0,477621 & 446 & 0,467 & 0,000 \\
    \cline{4-9} \cline{5-9} \cline{6-9}
     &  &  & Model trained only in the Landscapes's subset & 50 & 0,494020 & 414 & 0,532 & 0,000 \\
    \cline{4-9} \cline{5-9} \cline{6-9}
     &  &  & Model trained only in the Object's subset & 172 & 0,205238 & 159 & 0,543 & 0,000 \\
    \cline{4-9} \cline{5-9} \cline{6-9}
     &  &  & Model trained only in the People's subset & 141 & 0,305168 & 351 & 0,551 & 0,000 \\
     
    \cline{2-9} \cline{3-9} \cline{4-9} \cline{5-9} \cline{6-9}
     & \multirow[t]{19}{*}{2} & \multirow[t]{19}{*}{0,02} & General TF model trained in the Animal's subset & 146 & 0,490741 & 61307 & 0,461 & 0,000 \\
    \cline{4-9} \cline{5-9} \cline{6-9}
     &  &  & General TF model trained in the Landscape's subset & 108 & 0,488158 & 44508 & 0,527 & 0,000 \\
    \cline{4-9} \cline{5-9} \cline{6-9}
     &  &  & General TF model trained in the Object's subset & 146 & 0,491368 & 44990 & 0,537 & 0,000 \\
    \cline{4-9} \cline{5-9} \cline{6-9}
     &  &  & General TF model trained in the People's subset & 147 & 0,498451 & 61367 & 0,544 & 0,000 \\
    \cline{4-9} \cline{5-9} \cline{6-9}
     &  &  & General model trained in General dataset & 146 & 0,490741 & 61307 & 0,506 & 0,000 \\
    \cline{4-9} \cline{5-9} \cline{6-9}
     &  &  & Model trained only in the Animal's subset & 142 & 0,519296 & 585717 & 0,459 & 0,000 \\
    \cline{4-9} \cline{5-9} \cline{6-9}
     &  &  & Model trained only in the Landscapes's subset & 136 & 0,387359 & 4199259 & 0,517 & 0,000 \\
    \cline{4-9} \cline{5-9} \cline{6-9}
     &  &  & Model trained only in the Object's subset & 162 & 0,408242 & 45775 & 0,534 & 0,000 \\
    \cline{4-9} \cline{5-9} \cline{6-9}
     &  &  & Model trained only in the People's subset & 173 & 0,497688 & 8450039 & 0,542 & 0,000 \\
    \cline{1-9} \cline{2-9} \cline{3-9} \cline{4-9} \cline{5-9} \cline{6-9}
    \cline{1-9} \cline{2-9} \cline{3-9} \cline{4-9} \cline{5-9} \cline{6-9}
    \bottomrule
    \end{tabular}
    \end{table}

As demonstrated in Table \ref{tab10:resultados_comparacao_pso_ca_dsc_0.02}, the generalized model and the specialized models in the animal and object categories with $r = 2$ yielded closer values for each model parameter, indicating that the particle might have been stuck during the optimization phase, hindering its ability to effectively explore the search space to find better rules. Additionally, the proportion of categories in the dataset could have influenced this result. The animal and people categories collectively comprise 39\% and 30\% of the dataset, respectively.

Evaluating the adaptability of the models, the animal and object models demonstrated superior detection capabilities. They outperformed the general model in both their own category and the remaining categories, as evidenced by the figures \ref{fig:pso-ca-r1xr2-categories-psnr-0.02} and \ref{fig:pso-ca-r1xr2-categories-ssim-0.02}.

\begin{figure}[!htbp]
    \centering    
    \includegraphics[width=\textwidth]{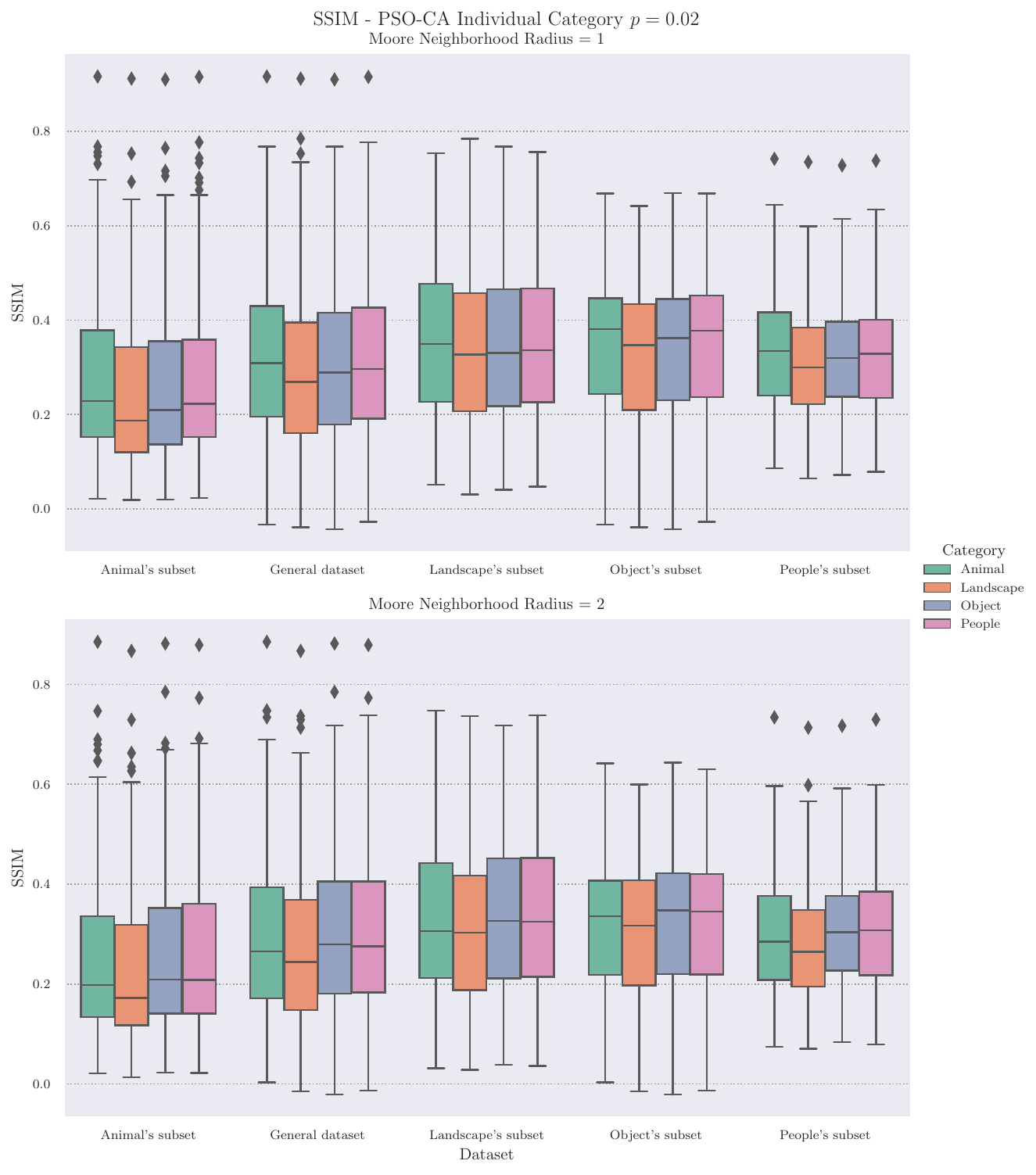}
    \caption{\textit{SSIM}}
    \label{fig:pso-ca-r1xr2-individual-categories-ssim-0.02}
\end{figure}

\begin{figure}[!htbp]
    \centering    
    \includegraphics[width=\textwidth]{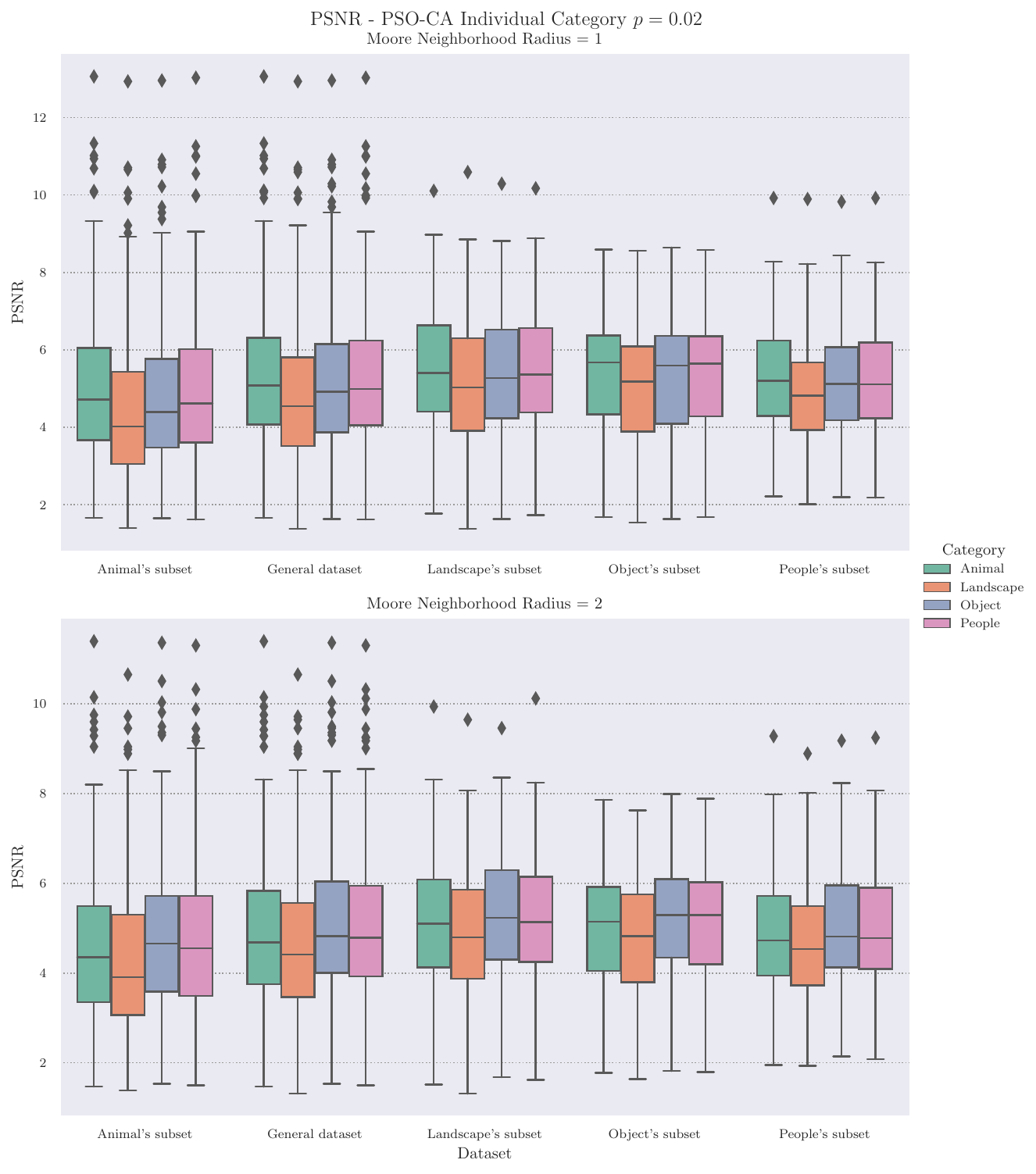}
    \caption{\textit{PSNR}}
    \label{fig:pso-ca-r1xr2-individual-categories-psnr-0.02}
    \caption{Comparison of the values of the \textit{PSNR} and \textit{SSIM} metric for the case study \ref{r_individual}. For each value of $r$ and each category of the database, the models trained exclusively in the categories are compared with the models trained in other categories.}
\end{figure}

The case study \ref{r_individual} (Figures \ref{fig:pso-ca-r1xr2-individual-categories-ssim-0.02} and \ref{fig:pso-ca-r1xr2-individual-categories-psnr-0.02}) presents the evaluation of the experiments of the individual models that don't employ Transfer Learning and are optimized solely using their own category dataset.

From the results found, even without the transfer of knowledge of the common domain and the presence of less training data, the models were also able to adapt to the characteristics of the images (Table \ref{tab10:resultados_comparacao_pso_ca_dsc_0.02}), presenting good results of \textit{PSNR} and \textit{SSIM} (Table \ref{tab9:resultados_comparacao_pso_ca_0.02}). For this experiment, the animal model presented the best results in both metrics for all subsets of the dataset, while the model specialized in the landscape subset presented the worst results.

\begin{figure}[!htbp]
    \centering
    \includegraphics[width=\textwidth]{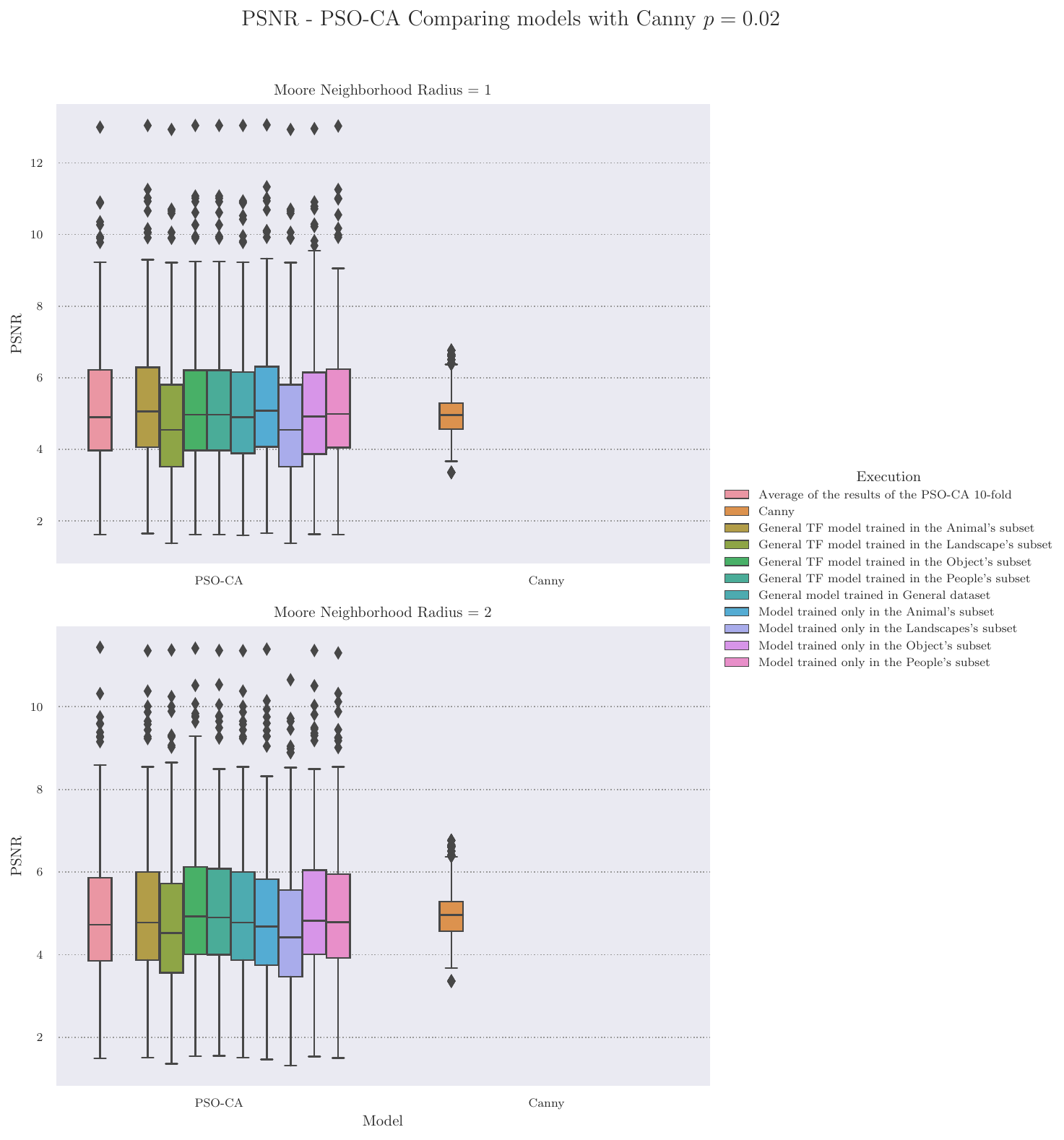}
        \caption{\textit{PSNR}}
        \label{fig:pso-ca-r1xr2xcanny-comparision-psnr-0.02}
\end{figure}

Similar to the specialized models, the models with $r = 1$ also showed better overall results and were able to perform better than Canny regardless of the radius and training category of the model (figure \ref{fig:pso-ca-r1xr2xcanny-comparision-psnr-0.02}).

An interesting remark of this experiment is that the models evaluated in it had closer values for the metrics when compared to the experiment \ref{r1_tf} and \ref{r2_tf}. Such behavior can be evidenced through the figures \ref{fig:pso-ca-r1xr2xcanny-comparision-ssim-0.02} and \ref{fig:pso-ca-r1xr2xcanny-comparision-psnr-0.02} and table \ref{tab9:resultados_comparacao_pso_ca_0.02}.

The specialized models, the ones enhanced with the Transfer Learning techniques, when compared to the individual models, showed fewer variations (Figures \ref{fig:pso-ca-r1xr2xcanny-comparision-ssim-0.02} and \ref{fig:pso-ca-r1xr2xcanny-comparision-psnr-0.02} and Table \ref{tab9:resultados_comparacao_pso_ca_0.02}). The results of the optimization phase reaffirm this behavior, presenting closer values for the \textit{DSC} (Table \ref{tab10:resultados_comparacao_pso_ca_dsc_0.02}).

Another insight found is that specialized and individual models trained in the landscapes category presented the worst average results of the metrics for all categories for both values of $r$. Such results of the landscape models can be explained by the characteristics of the category. Landscapes tend to present edge maps with very different characteristics than those of faces, animals, and objects.

When using the complete dataset, the general model presented results close to the specialized models and managed to circumvent the problem found in the models that only used the landscape category in training (Table \ref{tab9:resultados_comparacao_pso_ca_0.02} and figures \ref{fig:pso-ca-r1xr2-categories-psnr-0.02} and \ref{fig:pso-ca-r1xr2xcanny-comparision-ssim-0.02}).

\subsection{Comparing the proposed model with Canny}
\begin{figure}[!htbb]
    \centering
    \includegraphics[width=\textwidth,scale=2]{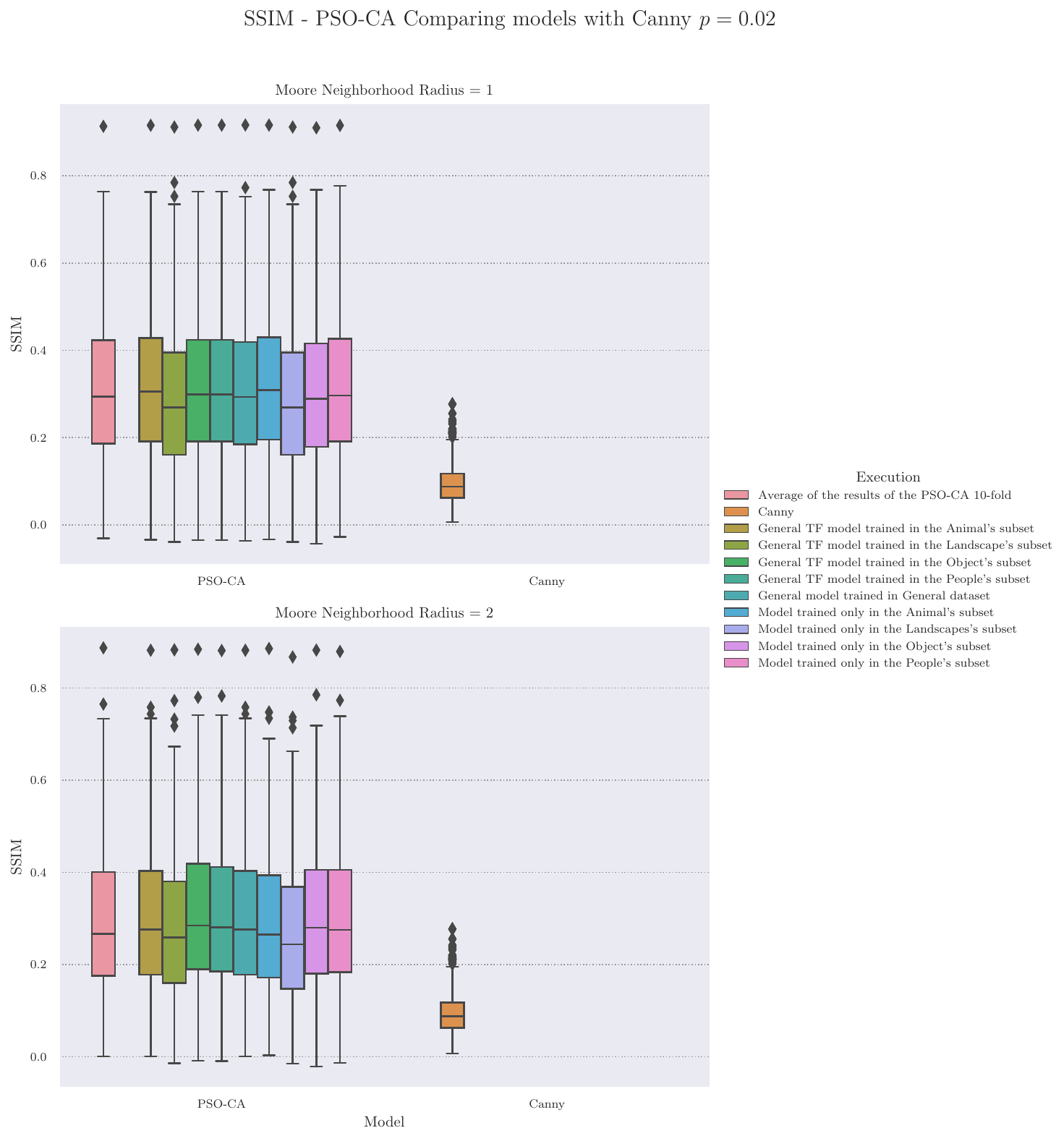}
    \caption{SSIM}
    \label{fig:pso-ca-r1xr2xcanny-comparision-ssim-0.02}
    \caption{Comparison of the values of \textit{SSIM} and \textit{PSNR} metrics between all case study models.}
\end{figure}

For all the experiments proposed in this study, the model studied was able to adhere to the properties of the image and present better results than Canny. For all cases, the \textit{SSIM} metric was the one that presented the best values for the proposed model and with the greatest differences when compared to Canny (Figure \ref{fig:pso-ca-r1xr2xcanny-comparision-ssim-0.02}).

The metric \textit{PSNR} also showed better results for the proposed model, but its difference was not so significant, and for some cases, presented results close to those of Canny (Figure \ref{fig:pso-ca-r1xr2xcanny-comparision-ssim-0.02}).

\section{Conclusions}
From the results, it is possible to conclude that for the scenario with $p = 0.02$, expanding the search space of the automaton rules does not bring benefits in the overall result since the model with $r = 1$ presented better or very close results in both metrics for most of the executions, indicating that the search space of the original model was already enough for these experiments.

The application of the Transfer Learning technique did not yield the expected adaptability. The results of the experiments demonstrate that it had a negligible impact on the specialization scenarios. The rules found for the general model and individual models (Figures \Ref{fig:pso-ca-r1xr2xcanny-comparision-psnr-0.02} and \ref{fig:pso-ca-r1xr2xcanny-comparision-ssim-0.02}) yielded close results. The proposed optimization phase, when combined with Transfer Learning techniques to explore the spaces of solutions of the automatons, proved ineffective. It was only able to find new rules with little variation of the fitness function.

The landscape model constantly presented the worst results for all variations of experiments. This result can be justified by the lack of intrinsic characteristics to it and the variation of landscape scenarios that, unlike humans, animals, and objects, may not have many things in common, such as similar features of faces and anatomies.

The results of the general model demonstrate that it was able to adapt to the dataset as a whole, presenting comparable results to specialized and individual models for most of the subcategories of images and better results for the subset of landscapes when compared to the specialized and individual landscape models (Figures \ref{fig:pso-ca-r1xr2xcanny-comparision-ssim-0.02} and \ref{fig:pso-ca-r1xr2xcanny-comparision-psnr-0.02}).

Cross-validation with 10 folds was also able to reinforce the adaptability of the model because, despite differences in training and separation of the sets, the final average results were very close for all experiments, and the values of the metrics found in the executions can be described by a common interval for the model (Figure \ref{fig:pso-caxcanny-fold-0.02-avg}).

When compared to the Canny detector, all variations of the proposed model were able to present better results in the evaluation metrics, confirming the adaptability of the model to the characteristics of image sets.

When analyzing the model with $r = 2$, it is possible that the granularity of the proposed \textit{PSO} domain, values in the range of 0 and 1, was not enough to represent the expansion of the search space of the rules accurately. Any small value of the range domain, when multiplied by the maximum number of the rule of $r = 2$, will present a rule number with little granularity, making it difficult to explore the search space in an accurate fashion.

Another topic of attention would be the optimization phase, when applying Transfer Learning techniques, the model did not present a better adherence to the characteristics of each category. This point can be justified by the values of \textit{DSC} and rules found in the optimization phase. Closer values indicate that the second stage of optimization for specialized models is getting stuck with particles and is not able to explore solutions with global values.

A proposition would be a variation of the parameters for the second stage of optimization to instigate the exploration of new solutions and a change in batch processing for both stages of optimization or only for the second stage. Changing the batch processing would help to decrease the dilution of the value of the \textit{DSC} which for the search is represented by the average of all images passed as input to the model.

For future studies, in addition to the considerations already mentioned, it would also be interesting to apply the proposed model to datasets with already predefined categories or a more robust process of pre-processing and division of categories such as the use of contrast metrics or other classifiers to help in impartiality, and the use of data augmentation techniques to balance the proportion of subcategories.

\bibliographystyle{abbrvnat} 
\bibliography{bibproj}

\end{document}